\documentclass[10pt,journal,compsoc]{IEEEtran}
\pdfoutput=1
\usepackage{hyperref}
\usepackage{amsfonts}
\usepackage{amsmath}

\usepackage{enumitem}
\usepackage{booktabs} 
\usepackage{graphicx}
\usepackage{subfigure}
\usepackage{tikz}
\usepackage{mathrsfs}
\usepackage{makecell}
\usepackage{multirow}
\usepackage[justification=centering]{caption}
\usepackage{float}

\usepackage{comment}

\usepackage{algorithmicx}
\usepackage[noend]{algpseudocode}

\usepackage{enumerate}

\usepackage[ruled]{algorithm2e} 

\newcommand{\stitle}[1]{\vspace{0.5ex}\noindent{\bf #1}}

\newcommand{\rev}[1]{ {{#1}}}
\newcommand{\kw}[1]{{\ensuremath {\mathsf{#1}}}\xspace}

\newcommand{\mae}{\kw{MAE}}
\newcommand{\rmse}{\kw{RMSE}}
\newcommand{\eat}[1]{}
\newcommand{\ie}{\emph{i.e.,}\xspace}
\newcommand{\eg}{\emph{e.g.,}\xspace}

\newcommand{\hisavg}{\kw{HA}}
\newcommand{\lsmrn}{\kw{LSM}-\kw{RN}}
\newcommand{\isbt}{\kw{ISBT}}
\newcommand{\graphsc}{\kw{GraphSC}}
\newcommand{\convlstm}{\kw{ConvLSTM}}
\newcommand{\stbfp}{\kw{RBFP}}
\newcommand{\ourmodel}{\kw{IBFP}}

\ifCLASSOPTIONcompsoc
  \usepackage[nocompress]{cite}
\else
  \usepackage{cite}
\fi
\ifCLASSINFOpdf
\else
\fi
\begin{document}
%

\title{Exploiting Interpretable Patterns for Flow Prediction in Dockless Bike Sharing Systems}
%
%
%
%

\author{Jingjing~Gu,
        Qiang~Zhou,
        Jingyuan~Yang,
        Yanchi~Liu,
        Fuzhen~Zhuang,
        Yanchao~Zhao,
        and~Hui~Xiong，~\IEEEmembership{Fellow,~IEEE}
\IEEEcompsocitemizethanks{
\IEEEcompsocthanksitem Jingjing Gu, Qiang Zhou and Yanchao Zhao are with the Department
of Computer Science and Technology, Nanjing University of Aeronautics and Astronautics, China.
E-mail: {gujingjing, zhouqnuaacs, yczhao}@nuaa.edu.cn. Jingyuan Yang are with George Mason University, USA. E-mail: jyang53@gmu.edu. Yanchi Liu and Hui Xiong are with Rutgers University, USA. {yanchi.liu, hxiong}@rutgers.edu.
 Fuzhen Zhuang is with the Key Laboratory of Intelligent Information Processing, Institute of Computing Technology, Chinese Academy of Sciences, China, with the University of Chinese Academy of Sciences, China,and also with the Henan Institutes of Advanced Technology, Zhengzhou University, China. E-mail: zhuangfuzhen@ict.ac.cn\protect\\

}
\thanks{Manuscript received April 19, 2005; revised August 26, 2015.}}

%
%

\markboth{Journal of \LaTeX\ Class Files,~Vol.~14, No.~8, August~2015}%
{Shell \MakeLowercase{\textit{et al.}}: Bare Demo of IEEEtran.cls for Computer Society Journals}
%



\IEEEtitleabstractindextext{%
\begin{abstract}
Unlike the traditional dock-based systems, dockless bike-sharing systems are more convenient for users in terms of flexibility. However, the flexibility of these dockless systems comes at the cost of management and operation complexity. Indeed, the imbalanced and dynamic use of bikes leads to mandatory rebalancing operations, which impose a critical need for effective bike traffic flow prediction. While efforts have been made in developing traffic flow prediction models, existing approaches lack interpretability, and thus have limited value in practical deployment. To this end, we propose an Interpretable Bike Flow Prediction (IBFP) framework, which can provide effective bike flow prediction with interpretable traffic patterns. Specifically, by dividing the urban area into regions according to flow density, we first model the spatio-temporal bike flows between regions with graph regularized sparse representation, where graph Laplacian is used as a smooth operator to preserve the commonalities of the periodic data structure.
Then, we extract traffic patterns from bike flows using subspace clustering with sparse representation to construct interpretable base matrices.  
Moreover, the bike flows can be predicted with the interpretable base matrices and learned parameters. 
Finally, experimental results on real-world data show the advantages of the IBFP method for flow prediction in dockless bike sharing systems. In addition, the interpretability of our flow pattern exploitation is further illustrated through a case study where IBFP provides valuable insights into bike flow analysis.

\end{abstract}

}

\maketitle

\IEEEdisplaynontitleabstractindextext

%
\IEEEpeerreviewmaketitle

\vspace{-3mm}
\section{Introduction}
The dockless bike sharing (DBS) system, serving as a new mode of public transportation, offers users a green and convenient way to travel. With the DBS systems, a user can pick up a bike using a mobile app, and drop off it anytime and anywhere without seeking an available docking station. Due to this flexibility, the bike sharing market has attracted a large number of users and has been growing at an amazing pace. \rev{The number of shared bike users has been up to 235 million, and 30 million new users are expected by 2019 \footnote{https://iimedia.cn}.} 
For example, Mobike (mobike.com), one of the largest DBS companies in the world, has launched bike sharing services in more than 200 cities all over the world, with more than 30 million rides each day.

\begin{figure}[t]
	\vspace{-5mm}
	\centering
	\includegraphics[width=0.6\linewidth,angle=90, bb=0 0 570 842]{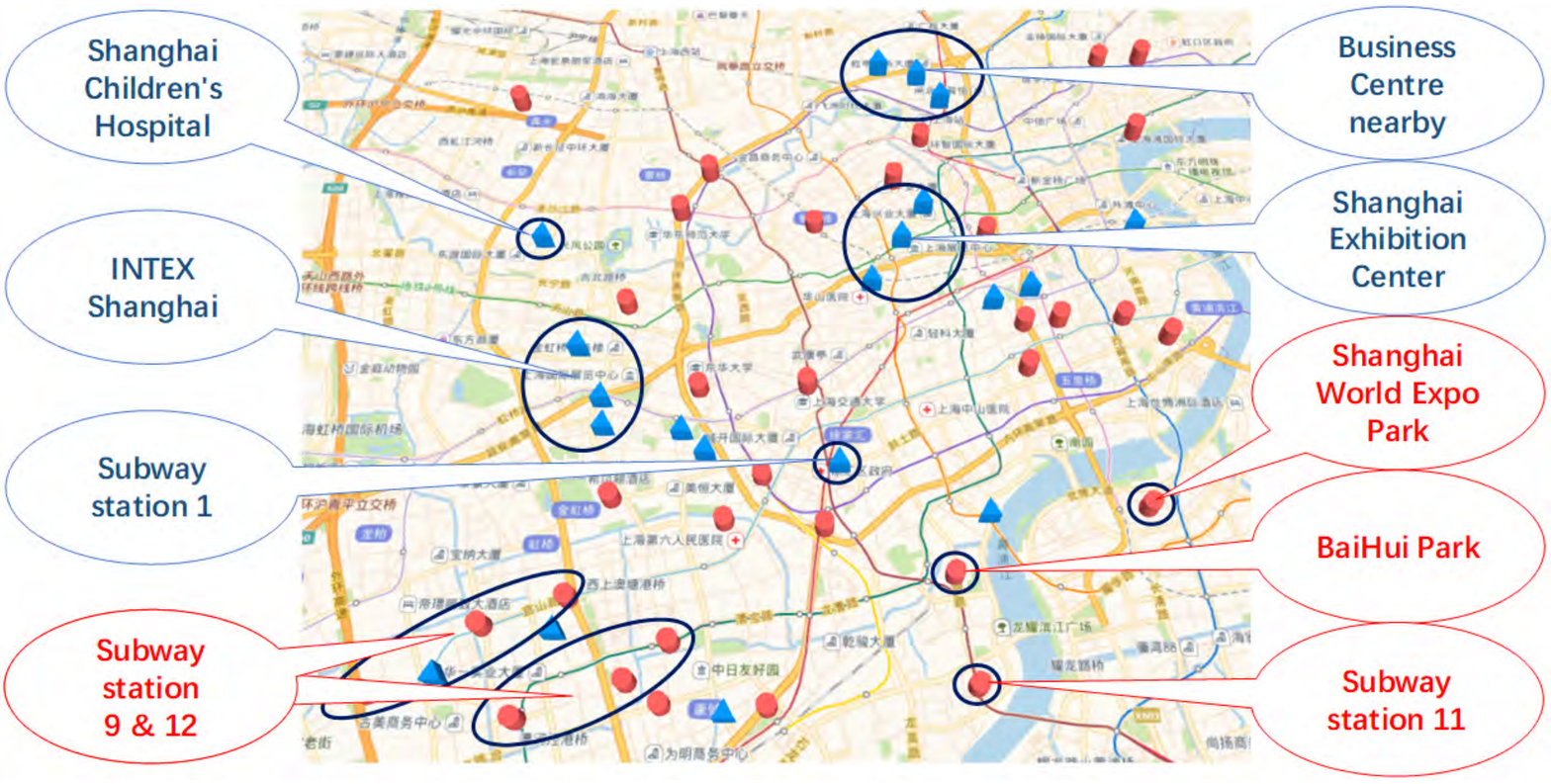}
	\vspace{-8mm}
	\caption{Imbalanced distribution of shared bikes: locations with high redundancy (red) and severe shortage (blue).}
	\vspace{-6mm}
	\label{fig:intro}
\end{figure}

Such large GPS-equipped bike sharing systems not only offer users a convenient way to travel, but also enable the new paradigm for the city management by providing ubiquitous sensing capabilities for understanding the city dynamics, such as traffic flows and the city-wide travel patterns. 
However, in DBS systems, a major challenge is the imbalanced bike distribution due to the uneven distributions of Points-of-Interests (POIs) and population, which impose tremendous management and operation challenges for both companies and governments.
\autoref{fig:intro} shows different locations in Shanghai with high redundancy (in red dots) and severe shortage (in blue triangles) of shared bikes, whose average hourly redundancy ratios (proportion of bikes which are not claimed in one hour) are more than 50\% and less than 3\%, respectively. The scope of the concerned site is the most popular area in Shanghai from our data, with the longitude from 121.35 to 121.5 latitude from 31.15 to 31.27. Each dot may cover one or a few combined grids ( 50m  $*$ 50m per grid).
The red locations with high redundancy ratios are mainly in the neighborhoods of museums and parks, such as the Shanghai World Expo Park. While the blue ones with low redundancy ratios are mainly near large commercial centers and hospitals. 
Moreover, it is interesting to see that, in some densely populated areas which usually have high bike usage demands (e.g. zones near different exits of the same subway station (9 and 12)), the redundancy ratios are quite different. Indeed, the imbalanced and dynamic use of shared bikes requires a constant and mandatory rebalancing operations~\cite{liu2016rebalancing}, and imposes a critical demand for effective bike traffic flow prediction.

There are some traditional approaches, such as regression \cite{singhvi2015predicting,li2015traffic} and time series analysis \cite{chemla2011balancing,deng2016latent} for flow prediction. However, these approaches do not consider the specific characteristics of bike flows such as the short riding distance. They have made some efforts on building more robust models with multi-source data and clustering techniques~\cite{liu2017functional,liu2016rebalancing} for bike flow prediction. Also, other researchers tried to use deep learning models to predict flows throughout a spatio-temporal network~\cite{zhang2019flow,li2018dynamic}. Nonetheless, they are mainly designed for dock-based shared bikes, and thus cannot be directly applied in dockless bike-sharing systems.  Recently, some works have been proposed based on the (graph) convolutional neural network \cite{lin2018predicting,geng2019spatiotemporal,chai2018bike,cui2019traffic,zhang2017deep}, which have excellent performances on flow prediction in scenarios for region-level flow (inbound and outbound flow). However, the (bike) flow between every OD (Origin to Destination) pairs were not studied. There are a few works about OD matrix prediciont \cite{deng2016latent,wang2019origin,gong2019potential}, but the scenarios they considered about are not closely relevant to ours.

Furthermore, it is also important to explore the traffic patterns to understand why and how the bike transfers within the city. Generally, some existing methods have incorporated the additional factors, such as rush hours, holidays, and weather conditions, as the constraints to regularize the optimization problems, and thus improve the prediction performances~\cite{liu2016rebalancing,hulot2018towards}. However, for this very flexible transportation mode which is available anytime and anywhere, it is difficult to describe those variables in a simple and consistent way. For example, intuitively, the bike traffic will decrease on rainy days. However, with observations from the real-world data, the bike use sometimes increase slightly during light rains. A possible reason is that, compared to walking, users can go fast with bikes, and thus suffer less in the rain. Moreover, we can hardly incorporate every factor in a fine-grained way, otherwise the model will become extremely complex and intractable.

To this end, in this paper, we propose an interpretable bike flow prediction framework (IBFP), which explores graph regularized sparse representation based on identified interpretable traffic patterns, for flow prediction in DBS systems.
Specifically, \textbf{(1)} we partition the whole city into regions according to different flow distribution densities,  rather than using traditional road segmentation and equal-size slicing, to avoid inaccessible regions and identify irregular regions. After this, bike flow matrices can be constructed. 
\textbf{(2)} We convert the problem of predicting whole flow matrices to the problem of predicting the components of flow matrices by modeling the matrix reconstruction problem. Here, we model the spatio-temporal interactions between region pairs with a graph regularized sparse representation, which characterizes the locally-invariant sparse of bike flow and preserve the commonalities of periodic and temporally related data structure by a graph smoothing operator.  
\rev{ \textbf{(3)} We cluster flow matrices to represent traffic patterns. Essentially, we characterize the high-dimension and sparse flow data by sparse representation, and use subspace clustering to infer the clustering of data through the sparse coefficients and similarity graph. Then we construct ``interpretable" base matrices from the corresponding patterns for being embed in the next prediction framework.  }
\textbf{(4)} The bike flows between region pairs can be predicted with the constructed interpretable base matrices and learned coefficients.

Finally, as a summary, the main contributions of this paper are as the following:

\begin{itemize}[leftmargin=8pt, topsep=0pt]
\item  We design a simple yet exquisite model to learn the traffic patterns in dockless bike sharing systems, which basically constructs interpretable base matrices to represent these patterns.
\item We propose a bike flow prediction method (IBFP) for the flow between each pair of OD regions in dockless bike sharing systems. IBFP exploits graph regularized sparse representation and graph smoothing factor to symbolize the flow attributes, and introduces a probabilistic factor model to approximate the real bike flow distribution.
\item We conduct extensive experimental studies on real-world data, whose results show the effectiveness of IBFP for bike flow prediction. Moreover, we also present several case studies to further demonstrate the interpretability of our approach.
\end{itemize}

\vspace{-2mm}

\section{Problem Formulation and System Framework}

In this section, we first introduce region partition, a fundamental step for dockless bike flow prediction. We then define our bike flow prediction problem, and finally, provide an overview of our system framework.

 \begin{figure}
	\centering
	\includegraphics[width=\linewidth, bb=0 0 393 214]{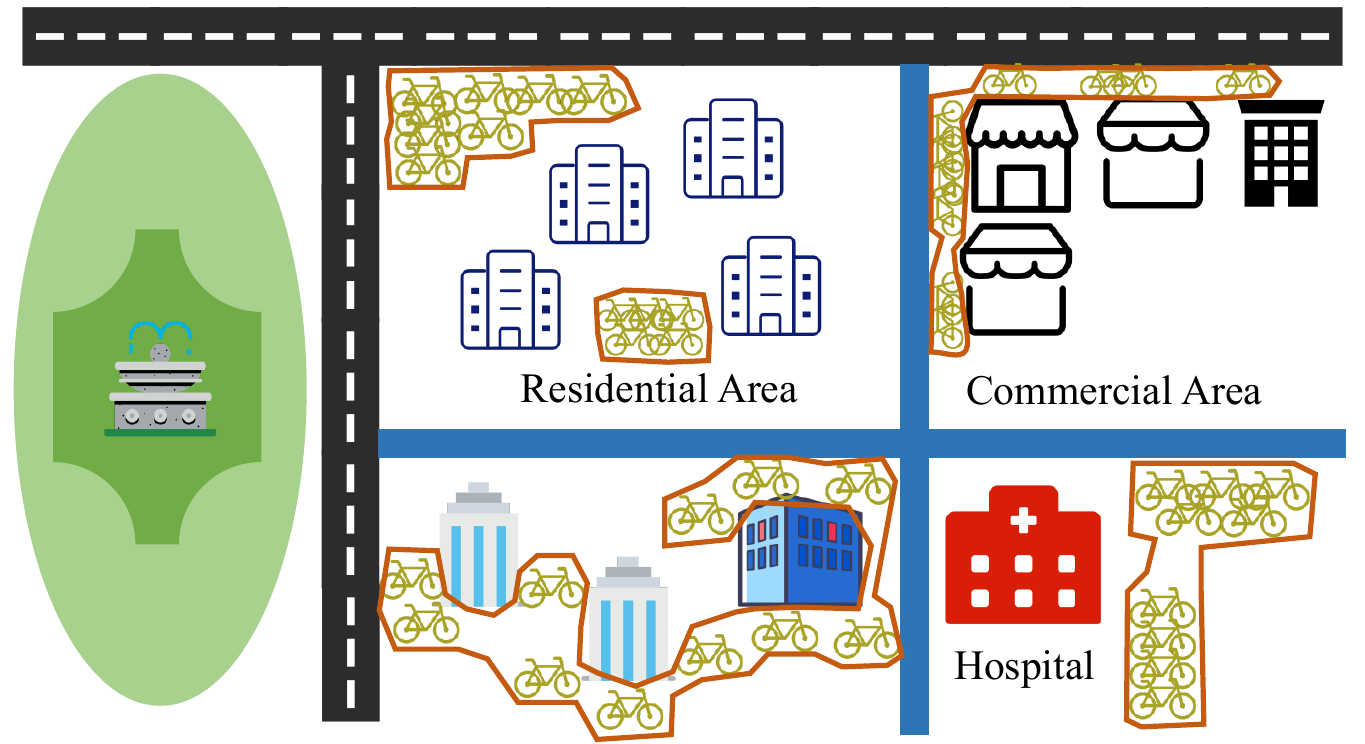}
	\vspace{-4mm}
	\caption{An example of the region partition.}
	\vspace{-4mm}
    \label{fig.dbscanexample}
\end{figure}

\vspace{-3mm}
\subsection{Region Partition} 
\label{subsec:region}


Without pre-established stations, DBSs calls for a more delicate formalization of the problem since bike flows between two points, \eg GPS locations, are hard and sometimes meaningless to predict. 
An illustration of specific region partition is shown in the \autoref{fig.dbscanexample}. As shown, the concerned area is already divided into regions by the road network. However, the partition is too coarse to precisely model bike flow. For example, there are two bike clusters in the residential area (up middle in the figure), which may correspond the different exits of the area. Note that the distribution and demand of the two cluster may differ and thus require finer-grained partition method. The situation is similar when using equal-size grids for region partition. On the other hand, the partition results of DBSCAN can handle the scenarios of inaccessible regions, irregular regions, and fine-grained regions well.
Inspired by these observations, we propose to cluster bike parking points into regions based on DBSCAN algorithm~\cite{erman2006traffic}, which is a classic density-based clustering method.
To achieve a meaningful region partition, it works well with detection of clusters that have irregular shapes and can easily avoid those forbidden areas. 

We first choose an arbitrary bike as a starting point and find all its surrounding bikes within a fixed distance (referred to as $epsilon$). For the starting point, if the number of surrounding bikes is less than a threshold (referred to as $minPts$), it is temporally treated as an outlier and gets skipped. Otherwise, it forms a cluster which includes all its surrounding bikes and itself. All of these surrounding bikes will perform the same expansion process until no bikes can be further included. We then repeat the above process by choosing a new bike and stop the process if all bikes are considered. Finally, all bikes marked as outliers are removed and regions are formed by detecting the boundaries of clusters. 
In this way, a key concept of the DBSCAN algorithm is core points which have at least $minPts$ points within distance $epsilon$. 
The $minPts$ is the minimum number of points required to form a dense region. Therefore, the higher $minPts$ is, the less regions will be generated. Besides, $\epsilon$ is the distance for searching surrounding bikes and is proportional to the size of regions. The detailed discussion will be given in the Section 4.4 ``Parameters Sensitivity". 

\vspace{-4mm}
\subsection{Problem Formulation}
\label{subsec:problem}

\begin{figure}[!t]
	\vspace{-4mm}
	\centering
	\subfigure[Flow matrix]{
		\includegraphics[width=0.6\linewidth, bb=0 0 791 329]{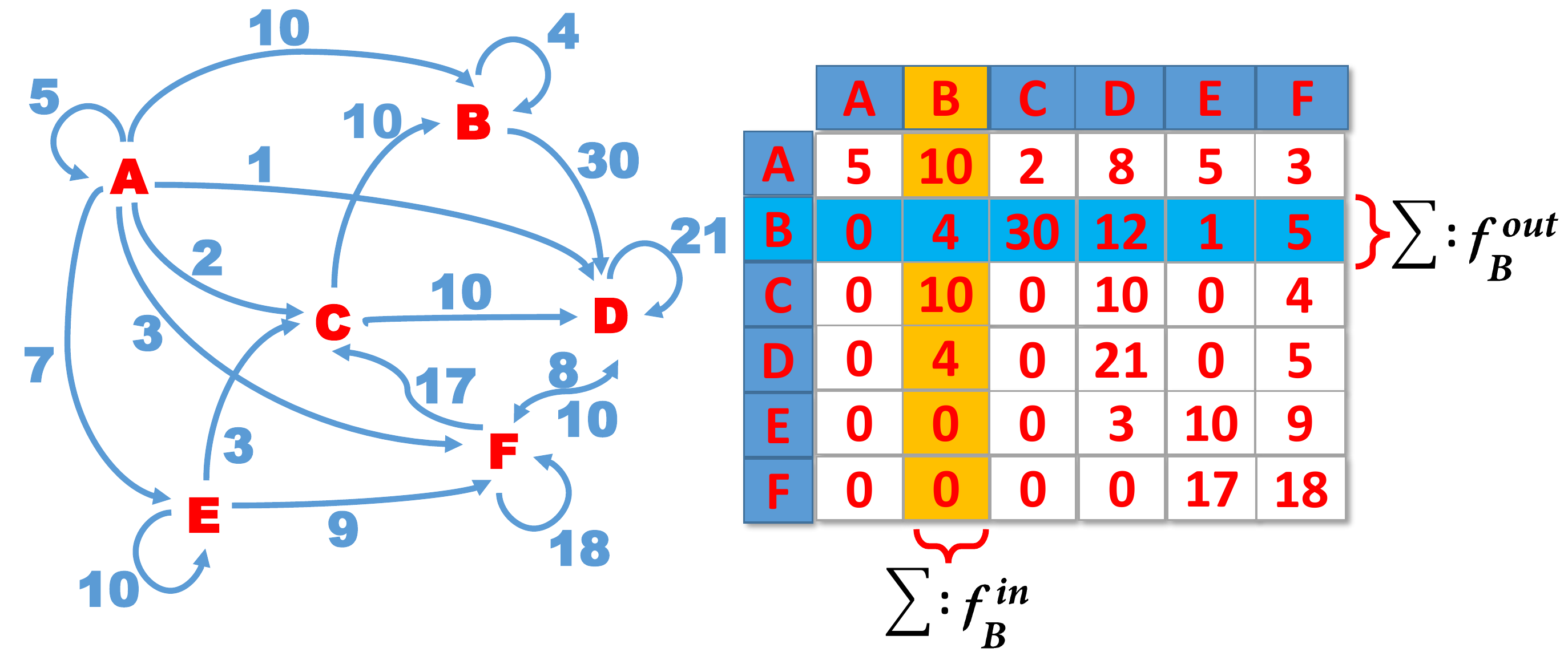}}
	\subfigure[A temporal series of flow matrices)]{
		\includegraphics[width=0.3\linewidth, height=62pt, bb=0 0 567 375]{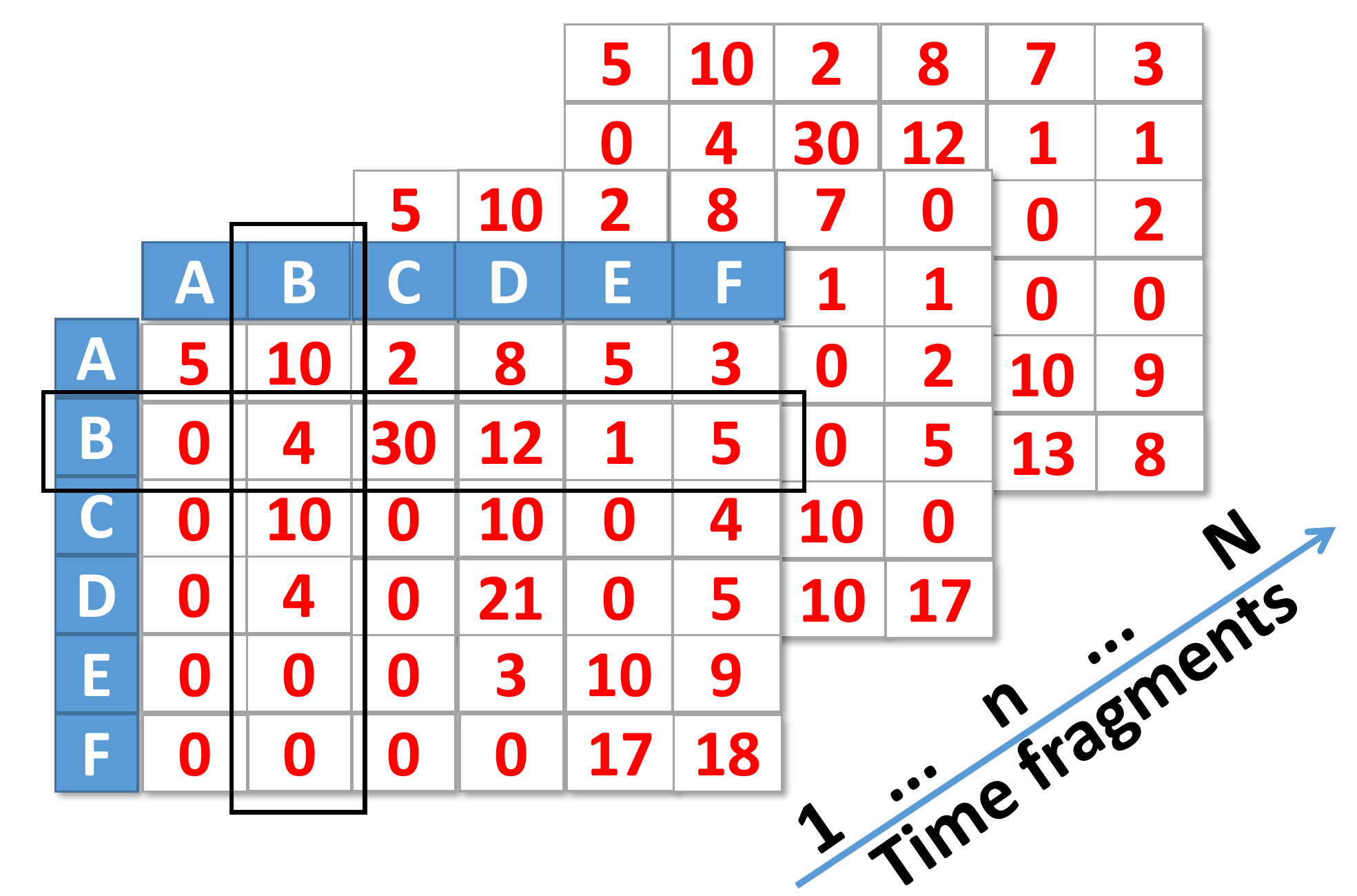}}
	\vspace{-4mm}
	\caption{An example of bike flow matrix construction}
	\vspace{-5mm}
	\label{fig:Matrix}
\end{figure}

After partitioning the entire city, a bike {\em flow matrix} can be constructed which records bike flows between any two regions in one time fragment. More specifically, suppose we have a total of $M$ regions. For a flow matrix  $F^n=\{f_{ij}^n\} {\in} R^{M{\times} M}$ on the $n$-th time fragment, $f_{ij}^n$ denotes the number of rides from the $i$-th region to the $j$-th region within the time fragment $(i=1,{\cdots},M; j=1,{\cdots},M)$. 
Further considering a sequence of $N$ time fragments, we then have a sequence of $N$ flow matrices, which can construct a flow tensor $\mathbb{F}=\{F^{1} ,F^{2},{\ldots},F^{N}\}$ with dimensionality $N\times M\times M$. 
\autoref{fig:Matrix} illustrates an example of bike flow matrices construction. Specifically, \autoref{fig:Matrix}  (a) shows the flow matrix of 6 regions in one time fragment from the records of bike flows between any two regions.
Particularly, we can also have the total inflow or outflow of any the $i$ region ($f_{i}^{in}$ and $f_{i}^{out}$) from the sum of the corresponding column or a row. \autoref{fig:Matrix} (b) shows a temporal series of flow matrices in $N$ time fragments.

The DBS flow prediction problem can be formulated as follows:
given a set of historical bike flow matrices, we aim to predict consecutive $h$ flow matrices from future time fragment $t$: $\{{F}^{t},{F}^{t+1},\ldots, {F}^{t+h}\}$. 
Note that: (a) in this work we consider a time fragment to be one hour, (b) the historical time fragments may be temporally non-consecutive due to multiple practical reasons such as data missing, (c) we predict day-level bike flows ($h=24$) in the experimental study.

\vspace{-4mm}
\subsection{System Framework}

\begin{figure}
	\vspace{-3mm}
	\centering
	\includegraphics[width=0.9\linewidth, bb=0 0 737 559]{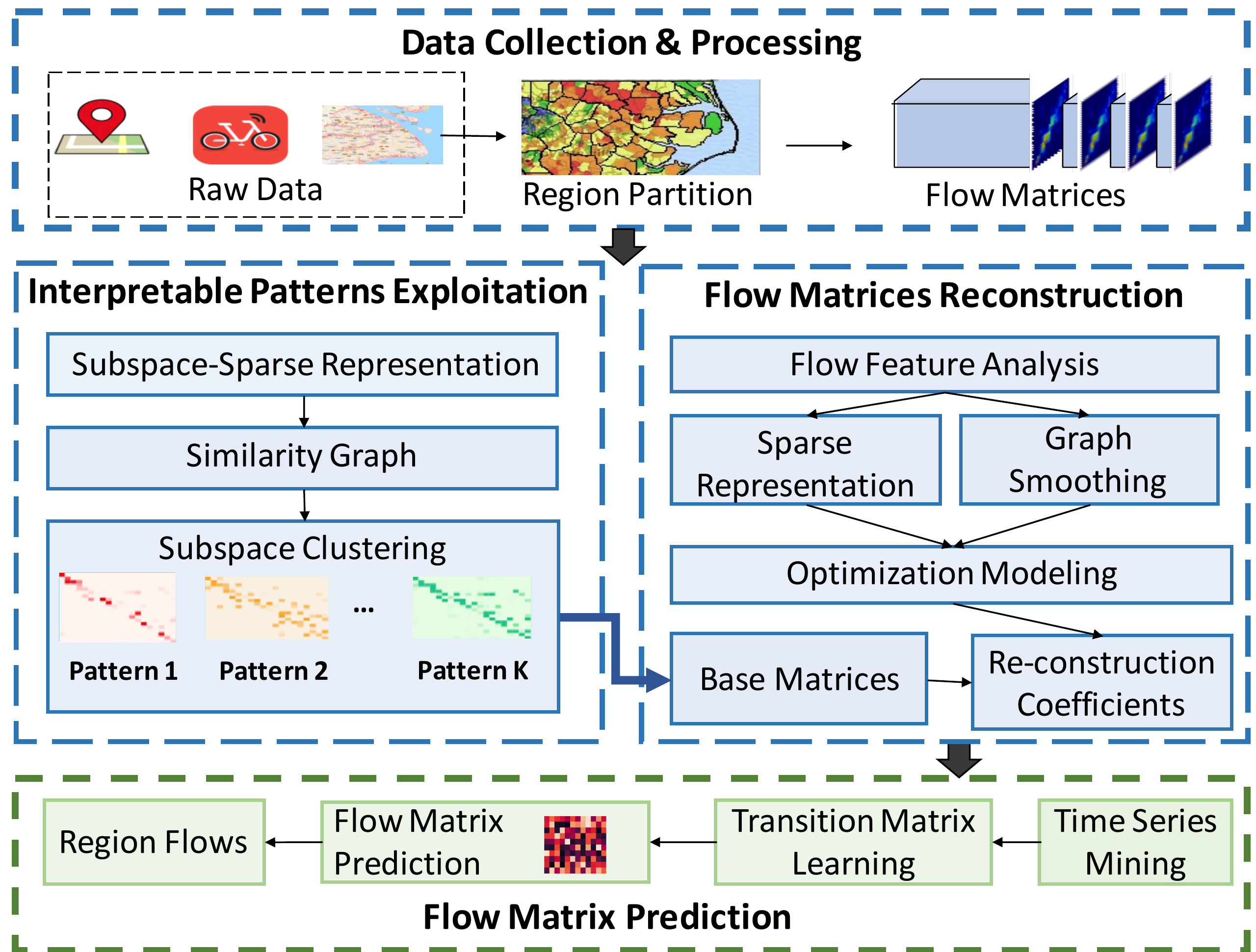}
	\vspace{-1mm}
	\caption{An Overview of the Framework.}
	\vspace{-4mm}
    \label{fig.Framework}
\end{figure}

As aforementioned, we formalize the bike traffic flow as matrices in time series. So basically, in our formulation, performing the flow prediction is equal to estimate flow matrices of future time slots. 
\autoref{fig.Framework}
shows the framework overview which consists of four major parts: data collection and processing, interpretable patterns exploitation, flow matrix reconstruction, and bike flow prediction.

\begin{itemize}[leftmargin=8pt, topsep=0pt]
\item {\textbf{Data collection and processing}.} 
We collect a large volume of bike riding records from Mobike, the largest shared bike companies in China. Each record specifies a bike ID, a pick-up location and time, a drop-off location and time. 
We first partition the city into regions with the location information in riding records, as described in Section \ref{subsec:region}, and then construct flow matrices by mapping each riding record into the corresponding time fragment and region pairs according to the spatio-temporal information, as described in Section \ref{subsec:problem}.

\item  {\textbf{Flow matrix reconstruction}.} 
To approximate the original flow matrices, we systematically specify the matrix reconstruction optimization problem by using the base matrices and the reconstruction coefficients. In the formulation, we notice that the flow matrices in DBS systems are exhibiting sparse and periodic structures. To solve this problem, we utilize sparse representation and graph smoothing to better capture these specific characteristics. After solving the reconstruction optimization problem, we can learn the reconstruction coefficients with the base matrices.

\item  {\textbf{Interpretable flow patterns exploitation}.} 
To take advantage of the potential traffic patterns, we cluster the existing flow matrices by utilizing a subspace clustering method with a sparse representation. After this, each cluster can be used to represent a hidden traffic pattern which can be used to construct interpretable base matrices for being embed in the prediction framework as a solution of the flow matrices reconstruction optimization.

\item {\textbf{Bike flow prediction}.} After the flow matrix reconstruction, in this step, we predict the future bike flows using a set of fixed bases and estimating the future reconstruction coefficients. The coefficients are mainly obtained by learning with transition matrix from previous transitions, which captures the historical temporal information and features of periodic evolving patterns.

\end{itemize}

\section{Interpretable Bike Flow Prediction}
As aforementioned, the key issue of our prediction approach is equivalent to predicting the flow matrix. Here, we first talk about how to reconstruct the flow matrix in section \ref{reconstruction}. Along with sparse representation and graph smoothing to characterize the bike flow features, and formulate an overall reconstruction optimization problems, whose the solution are base matrices and reconstruct coefficients.
Then, in section \ref{sec_bases}, we introduce extracting the bike flow patterns to construct our interpretable base matrices. 
Third, in section \ref{prediction}, we can further predict the flow matrix using the constructed base matrices and learned coefficients.
\vspace{-2mm}
\subsection{Flow Matrix Reconstruction}  \label{reconstruction}
\vspace{0mm}
 
Generally, a matrix can be expressed as a sum of a finite number of base matrices. Following this line, we compute a set of $C$ base matrices $\{ B^1,B^2,\ldots,B^C\} \in R^{M\times M}$ as the flow phenotypes, $C\ll N$. Then, let $S_{nc}$ be the reconstruction coefficient and $S\in R^{N\times C}$ be the coefficient matrix which is a representation for bike flow matrices of the $N$ time fragments.   Therefore, $B$ and $S$ can be used to reconstruct the observed flow matrix $F$.  
Moreover, $B$ and $S$ can be learned by approximating all the observed bike flow matrices with a minimized empirical loss function, which can be represented as $\frac{1}{2}\sum_{i=1}^{N}{\Vert F^i-\sum_{c=1}^{C}S_{ic}B^c\Vert}_F^2$, where $\Vert \cdot\Vert_F$ denotes the Frobenius norm.
The loss function for minimizing the reconstruction error can be further formulated as follows:
\begin{equation}
\setlength{\abovedisplayskip}{2pt}
\setlength{\belowdisplayskip}{2pt}
\label{eqn:01}
\mathcal{L}(B,S)=\frac{1}{2}\sum_{i=1}^{N}\Vert F^i-\sum_{c=1}^{C}S_{ic}B^c\Vert_F^2+\lambda\varphi(S),
\end{equation}
where $\lambda$ is a trade-off parameter and  $\varphi(S)$ is the function to measure the sparseness of $S$. A straightforward method to present the form of sparse constraints 
is the $\ell_1$-$norm$ regularization 
to approximate of the problem. Thus, we rewrite \autoref{eqn:01} and obtain:
\begin{equation}
\label{eqn:03}
\mathcal{L}(B,S)=\frac{1}{2}\sum_{i=1}^{N}\Vert F^i-\sum_{c=1}^{C}S_{ic}B^c\Vert_F^2+\lambda\sum_{i=1}^{N}\sum_{c=1}^{C}\Vert S_{ic}\Vert_1,
\end{equation}
where $\Vert S\Vert_1$  is a $\ell_1$-$norm$ which is convex and easy to be solved, and $\lambda$ controls the severity of the penalty. If $\lambda$ is too large, all parameters of the model will tend to be 0, resulting in under-fitting. On the other hand, if $\lambda$ is too small, it will have an over-fitting result.

The goal of minimizing the objective function \autoref{eqn:03} is trying to find a set of base matrices $\{ B^1,\ldots,B^C\}$ and a coefficient matrix $S$ to best approximate the original bike flow matrices $\mathbb{F}$.
One might further hope that we can use the intrinsic temporal relationships of the data structure. A natural assumption could be that if two bike flow matrices $F^i$ and $F^j$, where $i,j\in\{1,\ldots,N\}\&\;i\neq j$ from the same time interval and the same day of different weeks, they share similar characteristics of the data local structure.  For example, if $i$ and $j$ both represent the time fragment from 9:00 a.m. to 10:00 a.m. on Monday in different weeks, we may say that $F^i$ and $F^j$ tend to have similar inherent locality which possess a characteristic of manifold structures approximately. Therefore, we can use a smooth function to further discriminate periodic evolving patterns between flow matrices. Thus we let $W$ be the weight matrix of $\mathbb{F}$ and $W_{ij}=1$ if $F^i$ and $F^j$ in the same time intervals, otherwise $W_{ij}=0$. Apparently, $W$ is a the weighted graph and a symmetric matrix. We further utilize graph smoothing method  to preserve the periodic commonality of flow data in \autoref{eqn:03} and we have:
\begin{equation}
\setlength{\abovedisplayskip}{2pt}
\setlength{\belowdisplayskip}{2pt}
\begin{split}
\label{eqn:05}
\min\limits_{B,S}&\frac{1}{2}\sum_{i=1}^{N}\Vert F^i-\sum_{c=1}^{C}S_{ic}B^c\Vert_F^2+\lambda\sum_{i=1}^{N}\sum_{c=1}^{C}\Vert S_{ic}\Vert_1\\
&+\gamma\frac{1}{2}\sum_{i,j=1}^{N}\Vert S_i-S_j\Vert^2 W_{ij},
\end{split}
\end{equation}
where $\gamma\geq 0$ is the regularization parameter. 
\rev{

Moreover, let $L=D-W$ is a Laplacian matrix \cite{Belkin2002Laplacian}. $W_{ij}$ is symmetric and $D_{ii}=\sum_j W_{ij}$. 
For an arbitrary $c \in [1,C]$, $\frac{1}{2}\sum_{i,j=1}^{N}(S_{ic}-S_{jc})^2 W_{ij}$ can be written as:
\begin{equation}
\begin{split}
\label{eqn:added05}
&\frac{1}{2}\sum_{i,j=1}^{N}(S_{ic}-S_{jc})^2 W_{ij}
= S_{\cdot c}^TLS_{\cdot c}^{\ }.
\end{split}
\end{equation}
Therefore, given the $S=[S_{\cdot 1}S_{\cdot 2} \cdot \cdot \cdot S_{\cdot C}]$, we can have 
$\frac{1}{2}\sum_{i,j=1}^{N}\Vert S_i-S_j\Vert^2 W_{ij} = tr(S^TLS)$ \cite{Belkin2002Laplacian}.
}

By incorporating the constructed base matrices and Laplacian regularization into \autoref{eqn:05}, we can get the following objective function:
\begin{equation}
\begin{split}
\setlength{\abovedisplayskip}{2pt}
\setlength{\belowdisplayskip}{2pt}
\label{eqn:06}
\min\limits_{B,S}\frac{1}{2}\sum_{i=1}^{N}\Vert F^i-\sum_{k=1}^{K}S_{ic}B^k\Vert_F^2&+\lambda\sum_{i=1}^{N}\sum_{k=1}^{K}\Vert S_{ic}\Vert_1\\
&+\gamma\, tr(S^TLS).
\end{split}
\end{equation}

Now, the problem of predicting flow matrices can be converted to the problem of deriving the linear combination of $B$ and $S$, which are two solutions of the optimization problem in \autoref{eqn:06}. Generally, matrix reconstruction methods are mostly based on the assumption that the data is following the Gaussian distribution \cite{huang2018learning}, while this assumption does not hold for the bike flow data. Moreover, incomplete and biased observations \cite{mnih2008probabilistic} which may cause problems for accurately learning the above model.

 To solve this problem, we introduce a joint probability factor model to approximate the matrix reconstruction optimization based on the distribution of real flow data. Detailed formulations of the model are provided in Appendix A. Generally, the iterative process is composed of two steps: first learn the reconstruction coefficient matrix $S$ while fixing the base matrices ${B^1 ,B^2,{\ldots},B^C}$, and then learn these base matrices while fixing $S$ \cite{zheng2011graph}. 

However, this learning method is lacking of adequate interpretability. Also, satisfactory performance results may not be guaranteed due to that it is heavily affected by the random initialization. 
\rev{
To address these problems, we utilize the extracted traffic patterns described in Section \ref{sec_bases} to be embedded here. 
Consequently, we just need to learn the reconstruction coefficient of the optimization \ref{eqn:06} with the fixed ${B^1 ,B^2,{\ldots},B^C}$. 
\rev{
Detailed processes of the solution are also provided in Appendix A.
}
Finally, each flow matrix can be reconstructed via a weighted sum of these base matrices.
}

\vspace{-3mm}
\subsection{Extracting Bike Flow Patterns}\label{sec_bases} 

 \rev{
We are motivated to do so for the following reasons: (1) Traditional methods always consider the side factors in traffic flow with adding corresponding constraints to the optimization problem to regularize and match the target. This makes the model becomes extremely complex and the solution becomes intractable. (2) Bike flow naturally shares similar patterns and are jointly affected by multiple side factors (including weather, time, and location), which can hardly be modeled in a simple, regular, and empirical way. 
Accordingly, to model these side factors comprehensively, we cluster the flow matrices to extract features, and represent hidden traffic patterns using these clusters to construct the base matrices. }

\vspace{-2mm}
\subsubsection{Observations from Data Analysis}

\begin{figure}[!hth]
  \vspace{-4mm}
  \centering
  \subfigure[Distance distribution]{
    \includegraphics[width=0.23\textwidth, bb=0 0 595 372]{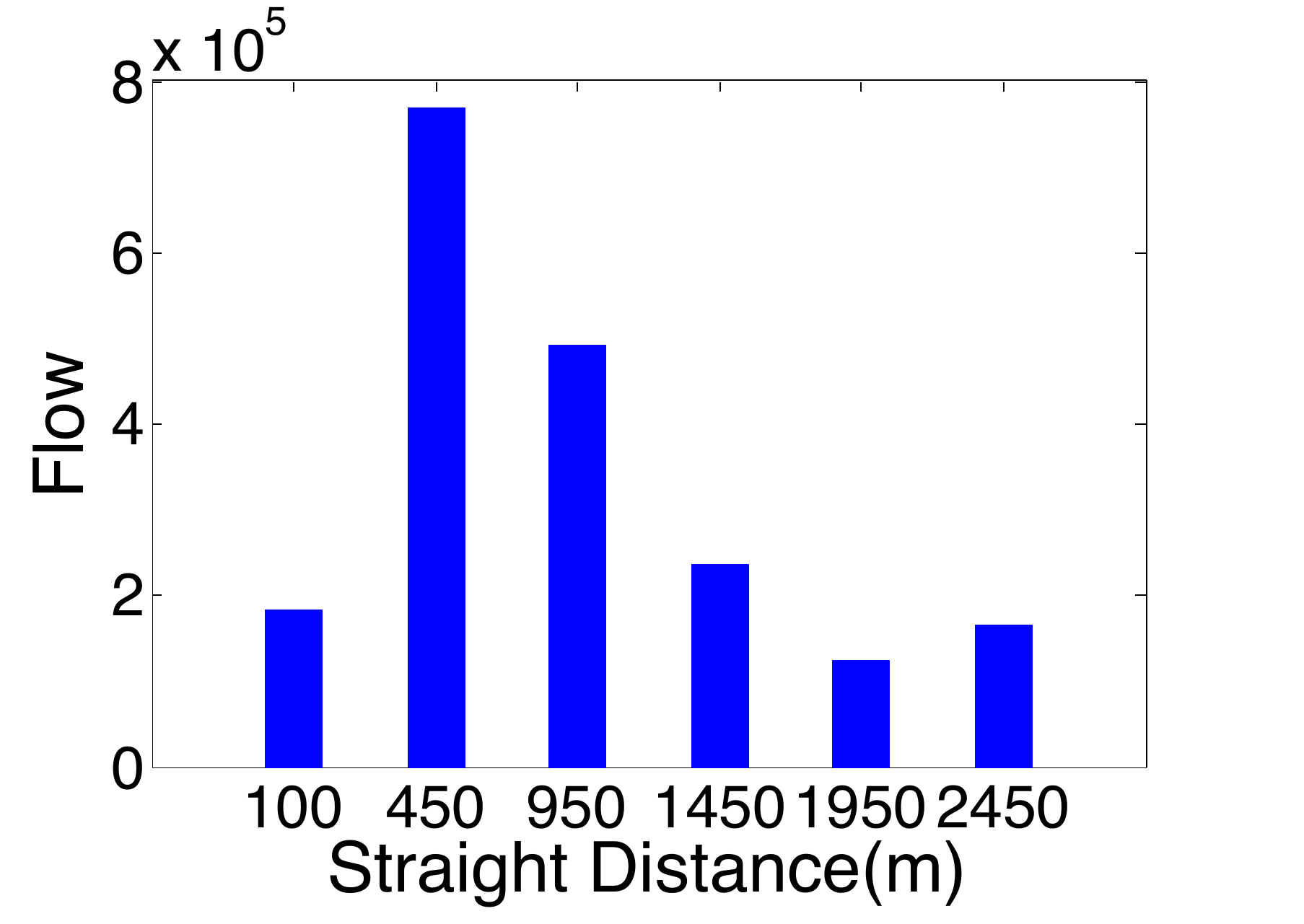}}
    \hspace{-4ex}
  \subfigure[Time duration distribution]{
    \includegraphics[width=0.23\textwidth, bb=0 0 595 372]{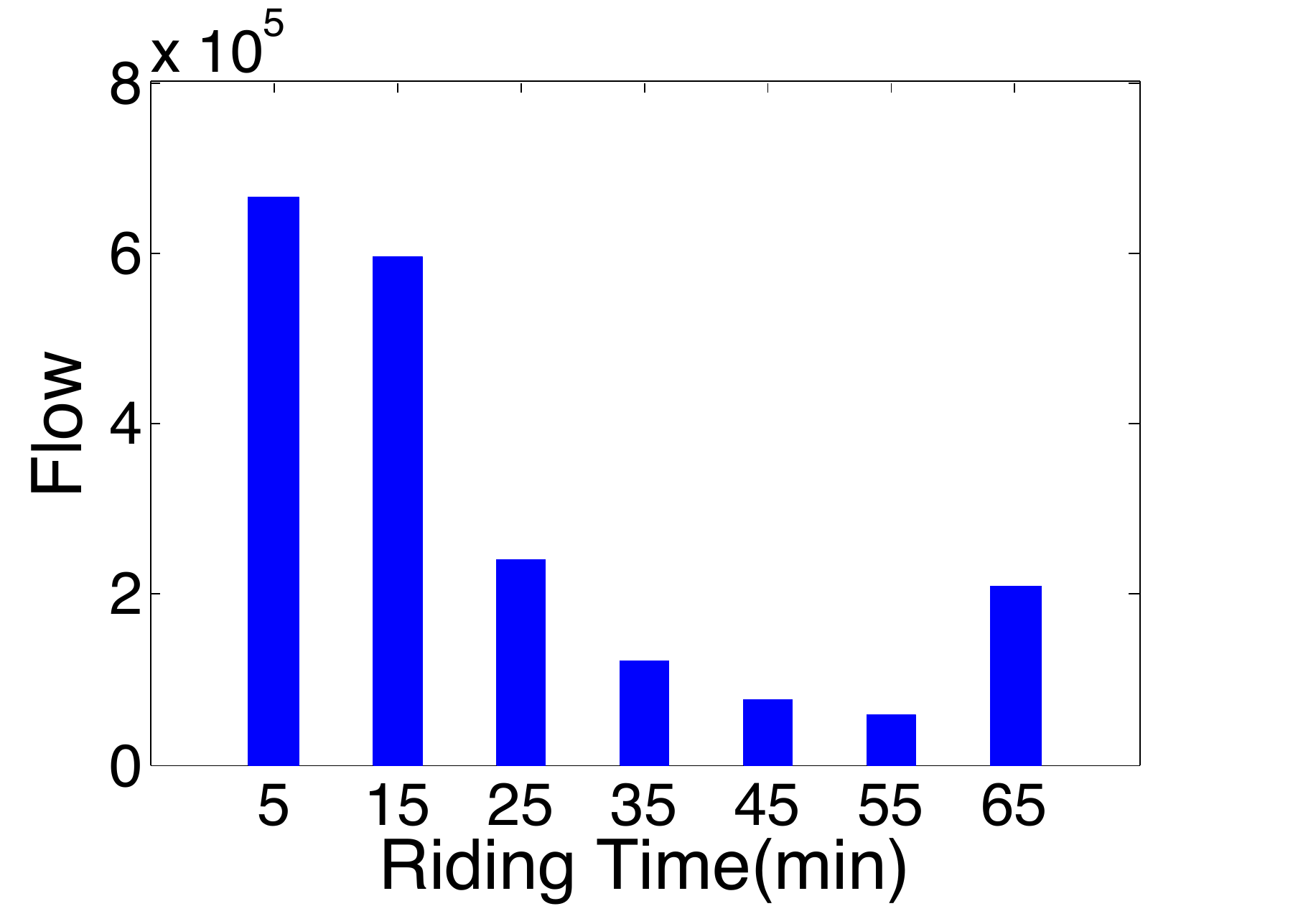}}
    \hspace{-4ex}
  \vspace{-4mm}
  \caption{Statistics of bike trips}
  \vspace{-3mm}
   \label{fig:stats}
 \end{figure}

We perform data analysis to understand the data characteristics.
Figure \ref{fig:stats} (a) and (b) show the statistical results of the straight distance and the time duration of bike trips in one week, respectively. The straight distance is the distance in a straight line from the pick-up location to the drop-off location of the shared bikes.
From this figure, we can observe that the flow matrix $F$ is sparse since most of the trips are between regions nearby. 
Another issue here is that the size of a flow matrix is as large as $766\times 766$ (as described in the section \ref{subsec:region}), which is not measurable by general similarity measurements, such as \emph{Euclidean distance}. 

To address these issues, we introduce the subspace clustering that seeks clusters in different sub-spaces, with the similarity graph based on the sparse representation.
More precisely, according to the \textit{self-expressiveness property} \cite{elhamifar2013sparse}, each data can be efficiently re-represented by a combination of all the other data in the dataset. Thus reconstructed sparse coefficients, whose nonzero elements correspond to the data from the same cluster (i.e. the same subspace), can be formulated as a similarity graph for clustering. 

\vspace{-2mm}
\subsubsection{Interpretable Base Matrices Construction}


By taking advantage of the ``self-expressiveness'' property of the data,
 each flow matrix can be written as: 
\begin{equation}
\vspace{-2mm}
  \label{eqn:self}
  F^{n}=\mathbb{F}\\\ c_n, c_{nn}=0,
\end{equation}
where 
$c_n = [c_{n1}, \dots , c_{nN}]^T$ and the constraint $c_{nn}=0$ eliminates the data that is expressed by itself. From the Equation \ref{eqn:self}, there exists a sparse solution $c_n$, whose nonzero entries $n$ correspond to data from the same subspace as the $F^n$. Thus, $c_n$ finds flow matrices from the same subspace where the number of the nonzero elements corresponds to the dimension of the underlying subspace.

A straightforward method to reduce the possible solutions is to minimize an objective function by the sparse constraints based on the $\ell_1$-$norm$ ~\cite{zheng2011graph}. 
Thus we can restrict the set of solutions as:
\begin{equation}
  \label{eqn:self_sparse}
\min {\parallel c_n \parallel}_1 \ \ \ \ \mathrm{s}.\mathrm{t}. \ F^{n} = \mathbb{F}\\\ c_n, c_{nn}=0
\end{equation}

We can also rewrite the sparse optimization program (\ref{eqn:self_sparse}) for all flow matrices $i = 1,...,N $ in the form as
\begin{equation}
  \label{eqn:self_sparse_all}
\min {\parallel C \parallel}_1 \ \ \ \ \mathrm{s}.\mathrm{t}. \ \mathbb{F} = \mathbb{F}\\\ C,
\\\ \mathrm{diag}(C)=0
\end{equation}
where $ C = [c_1, \dots , c_N] \in R^{N\times N} $ is the matrix whose $n-th$ column corresponds to the sparse representation of $F^{n}$. 
Equation \ref{eqn:self_sparse_all} can be solved efficiently using convex programming tools  \cite{elhamifar2013sparse}. 

After solving the proposed optimization program in (\ref{eqn:self_sparse_all}), we obtain a subspace-sparse representation for each flow matrix whose nonzero elements correspond to flow matrices from the same subspace, i.e. flow matrices that correspond to the same subspace are connected to each other and there are no connection between matrices in different sub-spaces. 

This provides us a way of building a similarity weighted graph $G=(V,\varepsilon, W^G)$, where $V$ and $\varepsilon$ denote the set of $N$ flow matrices and edges between flow matrices, respectively.
 $ W^G = (|C|+|C|^T), W^G \in R^{N\times N}$ is a symmetric non-negative similarity matrix.
In the $ W^G$, each node $F^{i}$ connects itself to a node $F^{j}$ by an edge whose weight is  $(|c_{ij}|+|c_{ji}|)$. Thus all flow matrices which correspond to nodes from the same subspace are connected to each other. Then, from $C$ connected components corresponding to the $C$ subspaces, we can have $W^G = diag \ ( W_1^G, \dots, W_K^G )$,
where $W_k^G$ is the similarity matrix in the subspace $S_k$. Then we can have a graph Laplacian matrix: $
  \label{eqn:Laplacianl}
L_w = I - D^{-1/2}W^G \  D^{-1/2},$
where $I$ is an unit matrix and $D$ is a diagonal matrix, where the $ith$ diagonal element is the sum of the $ith$ row of $W^G$. 
Compute the first $C$ eigenvectors $v_1, . . . , v_C$ of the $L_W$. 
We obtain the clusters of data by applying the K-means to the rows of the matrix whose columns are the C bottom eigenvectors of the $L_W$.

\vspace{-3mm}
\subsection{Bike Flow Prediction}\label{prediction}
In this sub-section, we study how to predict real-time flow from the learned parameters. Based on a set of fixed base matrices $\{B^1\cdots $ $B^C\}$ and reconstructed coefficients $S$ from our model described above, the flow matrix of the $n$-th time fragment can be reconstructed as: $F^i\approx \sum_{c=1}^{C}S_{ic}B^c$, where $S_i$ is the $i$-th coefficient vector in $S$ and $1\leq i\leq N$. Consequently, if we want to predict the flow of the further $t$-th time fragment: $F^{t}$, which can be reconstructed by $\sum_{c=1}^{C}S_{tc}B^c$, then we just need to calculate the coefficient vector $S_{t}$. Specifically, we can use a transition matrix $A_{t}\in R^{C\times C}$ to capture the evolving behavior from the $(t-1)$-th time to the $t$-th time and represent it as $S_{t}=S_{(t-1)}A_{t}$, where $S_{(t-1)}\in R^{1 \times C}$.
	
Generally, it is far from enough to use only the one flow information $F^{t-1}$ before the time fragment to predict $t$. However, it is impractical to make a prediction by using the whole training data in online prediction due to expensive computation cost. Moreover, it is not always true that more time information would yield a better prediction performance \cite{deng2016latent}. Inspired by this, we select some previous flow matrices which not only include the temporal information, but also can capture the inherent pattern information.
	
\begin{figure}[!th]
	\vspace{-5mm}
	\centering
	\includegraphics[width=1.0\linewidth, bb=0 0 1045 434]{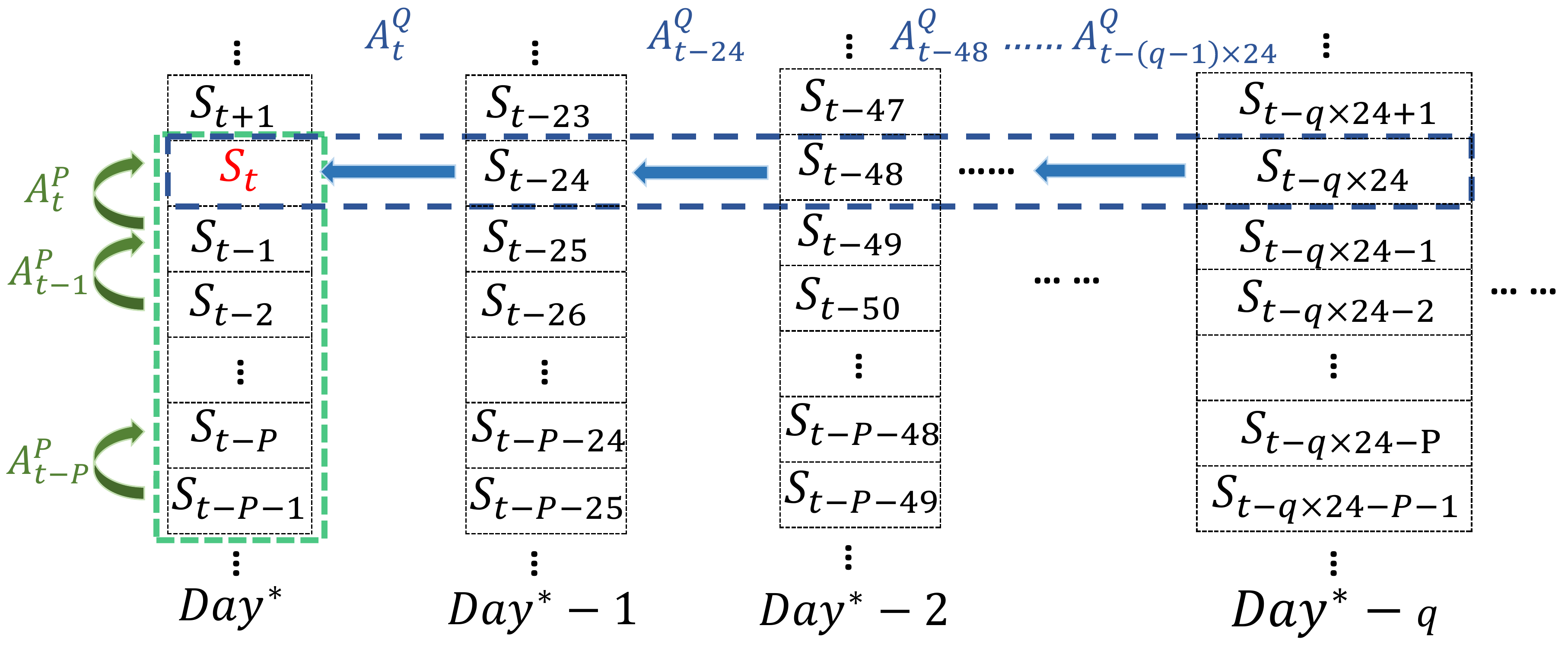}
	\vspace{-5mm}
	\caption{The prediction of coefficient from historical data}
	\vspace{-2mm}
    \label{fig:at}
\end{figure}

 Intuitively, there are some correlations between the behavior pattern of the predicting time and those of several previous times, which can be further divided into two kinds of time-related data for prediction, as shown in \autoref{fig:at},:
\begin{itemize}[leftmargin=4pt, topsep=0pt]
 \item {\bf Data with time fragments directly precedent the predicting time fragment.} We assume these historical data have some correlation with the predicting ones due to the continuous environmental pattern, such as temperature and air quality. Assume there are $p$ consecutive time fragments before the predicting one: $\{F^{t-1},\cdots$ $,F^{t-p}\}$, whose coefficient vectors are $\{S_{t-1},\cdots$ $,S_{t-p}\}$. Then we can have a sequence of transition matrices  $\mathcal{A}^P =\{A_{t-1}^P,\ A_{t-2}^P,\ \cdots$ $,\ A_{t-p}^P\}$, which forms a set of historical transitions:
 \end{itemize}  
 
  \noindent  \begin{equation}
   \{{ {S_{t-1}=S_{t-2}A_{t-1}^P};
   \ \cdots\;{S_{t-P}=S_{t-P-1}A_{t-P}^P}}\} 
     \end{equation}
     
\begin{itemize}[leftmargin=4pt, topsep=0pt]   
\item {\bf Data with time fragments repeating every day/week.} Such periodicity helps us to capture common features of periodic evolving patterns. For example, there is a correlation between 7 a.m. to 8 a.m. of working days. Subsequently, there is a transition from the previous day to the present day at a same time. Assume there are continuous $q$ days before the predicting time, thus we have a historical sequence of this fixing prediction time in varying $q$ days: $\{F^{t-24},\ F^{t-48}\cdots$ $,\ F^{t-{q \times 24}}\}$ under the condition of 24 time fragments a day. Similarly, we can also have a sequence of transition matrices  $\mathcal{A}^Q =\{A_{q-24}^Q,A_{q-48}^Q,\ \cdots$ $,\ A^Q_{t-q\times 24}\}$, which forms the corresponding set of historical transitions:
     \begin{equation}
     \begin{split}
    \{S_{q-24} =S_{q-48}A_{q-24}^Q;&
    \cdots ;\ \\
    &S_{t-(q\times 24)}
    =S_{t-(q+1) \times 24}A_{t-q\times 24}^Q\}
    \end{split}
    \end{equation}    
\end{itemize}    

  From the primarily experiments here, we find generally $p$ is in the  range of 3 to 6 and $q$ is in the range of 12 to 20, the prediction performance are stable. Note that, in our model, when we predict the weekday flow, we will omit the weekend flow data, as the flow features may have great differences between the weekdays and weekends. From $\mathcal{A}^P$ and $\mathcal{A}^Q$, we can predicted two transition matrices $A_{t}^P=\frac{1}{p}\sum_{i=1}^{p}\mathcal{A}_{i}^P$ and $A_{t}^Q=\frac{1}{q}\sum_{j=1}^{q}\mathcal{A}_{j}^Q$ with the similarities of the highest temporal duration and the periodic evolving pattern, respectively. By incorporating two transition matrices, we can have   
     \begin{equation}
	  S_{t}={\frac{1}{2}}\cdot( {S_{t-1} {\frac{1}{p}\cdot\sum_{i=1}^{p}\mathcal{A}_{i} }+S_{t-24}\cdot {\frac{1}{Q} \cdot\sum_{j=1}^{Q}\mathcal{A}_{j}})}\\
	\end{equation}
    
    Moreover, we can have an incremental method to predict the flow of the next $(t+h)$-th time. From $S_{t}=S_{t-1}A_{t}$, we can have $S_{t+1}=S_{t}A_{t+1}=(S_{t-1}A_{t})A_{t+1}$. Follow this line, we can have:
$
	 S_{t+h}=S_{t-1}\cdot \prod_{i=t}^{t+h}A_i.
$
   As a result, we can predict bike flows of the $(t+h)$-th time as:
     \begin{equation}
     \setlength{\abovedisplayskip}{2pt}
\setlength{\belowdisplayskip}{2pt}
	    F^{t+h}=\sum_{c=1}^{C}S_{t+h}B^c=\sum_{c=1}^{C}(S_{t-1}\cdot \prod_{i=t}^{t+h}A_i)B^c,
	\end{equation}
   where $h$ is the number of future time fragments. After predicting the flow matrix of the $(t+h)$-th time, we can further calculate the outflow and inflow of each region. As shown in \autoref{fig:Matrix}, $f_{ij}^n$ is the flow from region $i$ to $j$ at the $n$-th time fragment. Thus, all elements of the $i$-th row are flows with the departure region $i$ and all elements of the $j$-th column are flows with the destination region $j$. Thus, on the time segment $t$, the outflow $f_i^{out,t}$ and inflow $f_i^{in,t}$ of the region $i$ are represented as:
    \begin{equation*}
    \setlength{\abovedisplayskip}{2pt}
\setlength{\belowdisplayskip}{2pt}
       \label{eqn:00}
        f_i^{out,t}=\sum_{i=1}^{M}f_{:,i}^t \qquad f_i^{in,t}=\sum_{i=1}^{M}f_{i,:}^t 
      \end{equation*}

\newcommand{\gscale}{1.6}

\section{Experimental Results}

In this section, we present an extensive experimental study of our interpretable \ourmodel model, compared with six competitive algorithms. Using real-life data, we conduct a sets of experiments to evaluate the effectiveness of our \ourmodel model and study the parameter sensitivity. 
We also present an interesting case study as well as our findings regarding  the interpretability of bike flow prediction. The source code for IBFP is available at https://github.com/Enlindn-NUAA/IBFP.
\vspace{-2mm}
\subsection{The Experimental Setting}

We first introduce the settings of our experimental study.

\stitle{Experimental Data}. We collected the real-life Mobike data to test our model. It contains 957,357,367 riding records generated by 314,703 shared bikes from February 2017 to March 2018 in Shanghai city. Each record specifies a bike ID, a pick-up location, a pick-up time, a drop-off location, and a drop-off time. 
Since bike usage is sensitive to holidays and weather conditions, we treat the data on working days and holidays separately. The dataset spans 291 working days data and 132 holidays, respectively, among which 48 (16.5\%) working days and 22 (16.6\%) holidays are randomly selected as test set and the rest are used for formulating the training dataset. To investigate the impacts of weather conditions, the test set on working days is further split into two parts: one consisting of 35 regular working days and the other consisting of 13 rainy working days. We do not further split the holiday test data due to the limited number, \ie 5, of rainy days. To conclude, we will train our model based on two training sets, \ie working days and holidays, and validate our model with three test sets, \ie (1) regular working days, (2) rainy working days and (3) holidays.

\stitle{Baselines}. We compare our \ourmodel model with six algorithms.

\noindent (1) Historical Average (\hisavg)~\cite{kamarianakis2003forecasting} directly uses the average bike flow at the same day time in training data.

\noindent (2) Latent Space Model for Road Networks (\lsmrn)~\cite{deng2016latent} utilizes Nonnegative Matrix Factorization (NMF) for predicting speed and we replaced the speed matrix with flow matrix for flow prediction. 

\noindent (3) Inter Station Bike Transition (\isbt)~\cite{liu2016rebalancing} predicts station-based bike traffic flows by modeling bike usage demand and trip duration. In our dockless setting, it treats the regions detected by our IBFP model as stations. 

\noindent (4) Graph Regularized Sparse Coding (\graphsc)~\cite{zheng2011graph} is a state-of-the-art model for sparse representation.It uses the same prediction method as our IBFP model after the base matrices has been constructed.

\noindent (5) Convolutional Long Short-Term Memory(\convlstm) ~\cite{xingjian2015convolutional} extends the traditional fully connected LSTM (FC-LSTM). It has convolutional structures in both the input-to-state and state-to-state transitions, which will keep the spatial information.

\noindent (6) Random Bike Flow Predictions (\stbfp) is a simplified version of our IBFP model. It is the same to IBFP except that the base matrices are learned automatically without considering the traffic patterns.

\stitle{Metrics}.
We evaluate the performance by three metrics, namely mean absolute error (\mae)  and root mean square error (\rmse). Note that for both of the two metrics, a lower value indicates a better accuracy.


\stitle{Implementation}. The default parameters are selected as follows. (a) For all algorithms based on sparse representation, the number $C$ of base matrices is set to 9. (b) The dimensionality of the latent space is set to 10 for \lsmrn. (c) The parameters of \isbt, \convlstm and the regularization parameters of \graphsc are chose following\cite{liu2016rebalancing}, \cite{xingjian2015convolutional} and\cite{zheng2011graph}, respectively. (d) For \stbfp and \ourmodel, please see Section~\ref{exp-para}.

In our experiments, we have partitioned 766 irregular regions from DBSCAN, which is unsuitable for ConvLSTM. Thus, to make a fare comparison with ConvLSTM, we divided the same area into 768 (32 $\times$24) regular regions. In this way, the number of regions of all baselines and our method are almost same.
All experiments are run on a PC with a 3.6 GHz Intel(R) Core i7-4790 CPU, and 16 GB RAM running the 64-bit Windows 10 system. When quantity measures are evaluated, the tests are repeated over 10 times and the average results are reported.


\eat{
\begin{figure}[t]
	\vspace{-2mm}
	\centering
	\subfigure[ ]{
		\includegraphics[height=\gscale in]{./Fig_Overall_Datasets_ER.pdf}}
	\subfigure[ ]{
		\includegraphics[height=\gscale in]{./Fig_Overall_Datasets_MAE.pdf}}
	\subfigure[ ]{
		\includegraphics[height=\gscale in]{./Fig_Overall_Datasets_RMSE.pdf}}
	\vspace{-3mm}
	\caption{ Overall Dataset.  }
	\vspace{-2mm}
    \label{res_dataset1}
\end{figure}
}
\vspace{-4mm}

\subsection{A Performance Comparison}
In the first set of experiments, we compare the flow prediction errors of our \ourmodel model with the six algorithms on two test sets which are introduced above. 

 \textbf{(1) Effectiveness on regular working days.}
We first evaluate the effectiveness on regular working days when bike usage demands are concentrated on morning and evening peaks and people are willing to ride bikes. 
We test the prediction errors at different day time. The results are reported in \autoref{res_dataset1}.

For (almost) all algorithms, the prediction errors have two peaks at the morning and evening, which is due to the sudden increase of bike usage demands. Furthermore, performances of almost all algorithms are unexpectedly decreased at around 6 a.m. due to the inaccurate training data caused by official relocation of the bikes.

Moreover, our \ourmodel model consistently performs the best in all our tests. Indeed, compared with (\hisavg, \lsmrn, \isbt, \graphsc , \convlstm, \stbfp), \ourmodel decreases the errors 
both under \mae and \rmse. Our simplified \stbfp model already outperforms other approaches, which demonstrates the superiority of utilizing graph regularized sparse representation on flow matrices for bike flow prediction. Our \ourmodel model further outperforms \stbfp, indicating that the constructed base matrices are not only more interpretable but also more robust. 

\begin{figure}[t]
	\vspace{-2mm}
	\centering
	\subfigure{
		\includegraphics[width=0.23\textwidth, bb= 0 0 459 345]{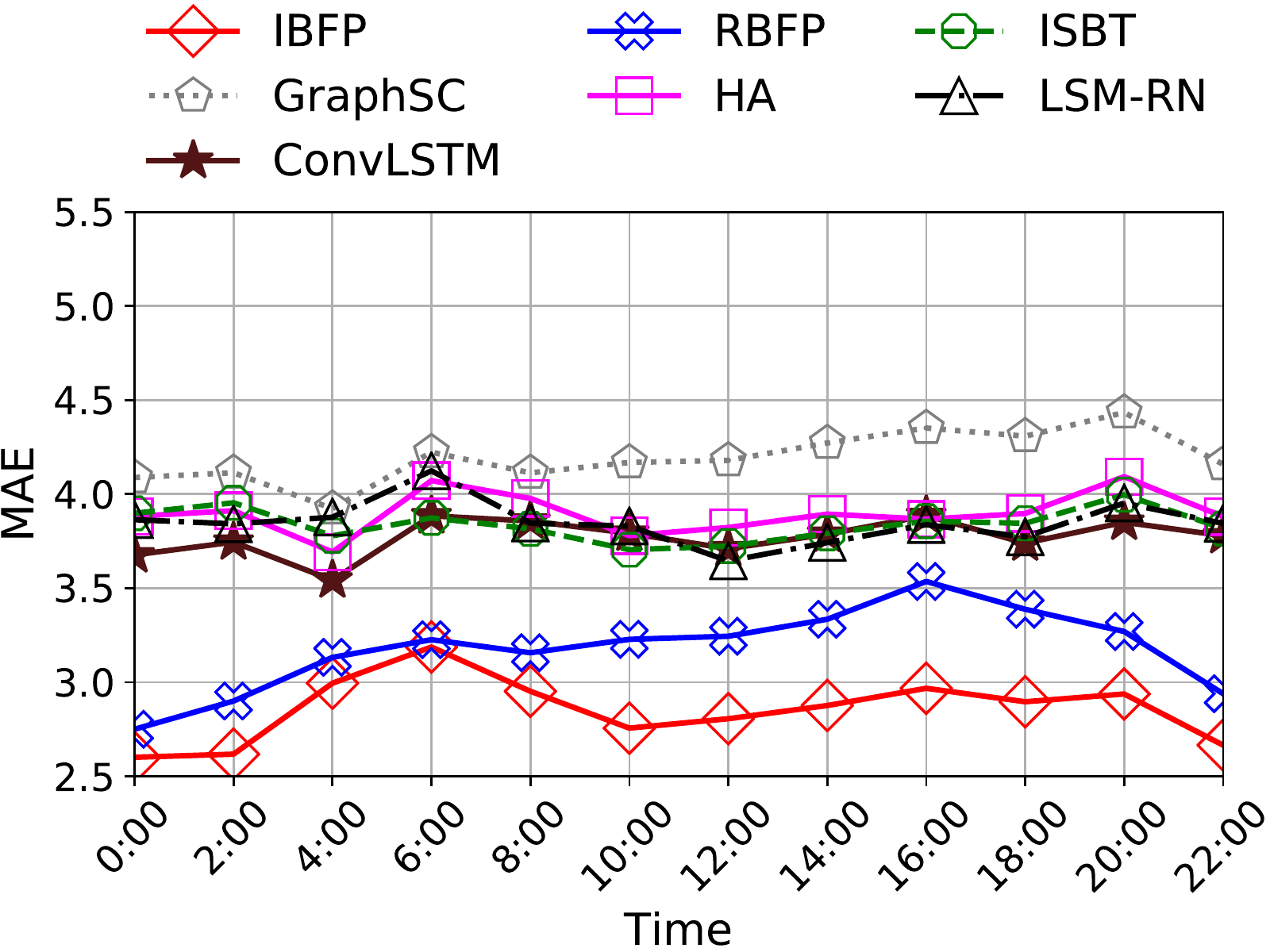}}
	\subfigure{
		\includegraphics[width=0.23\textwidth, bb= 0 0 459 345]{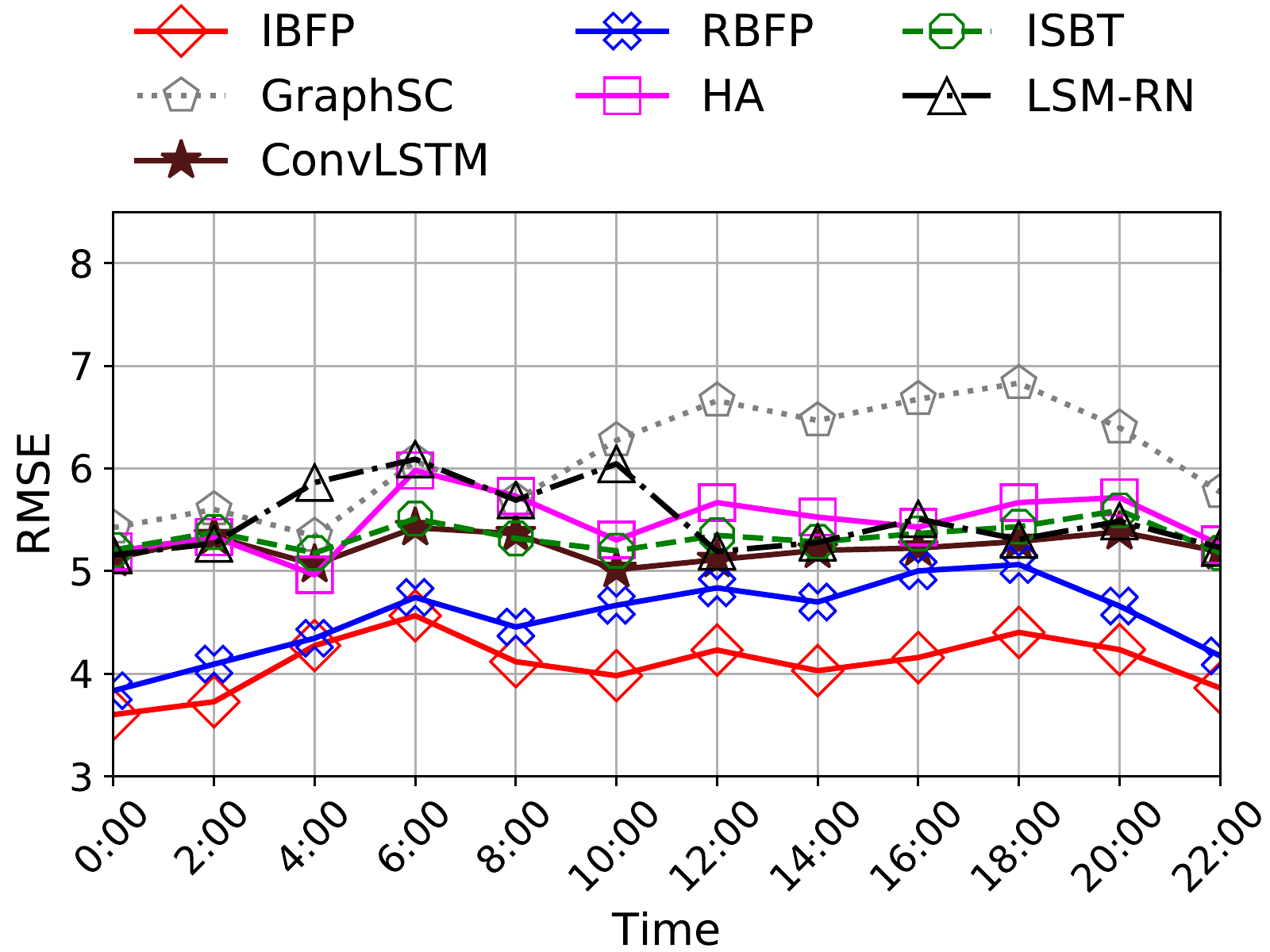}}
	\vspace{-3mm}
	\caption{Prediction errors on regular working days}
	\vspace{-3mm}
    \label{res_dataset1}
\end{figure}

\begin{figure}[t]
	\vspace{-2mm}
	\centering
	\subfigure{
		\includegraphics[width=0.23\textwidth, bb= 0 0 459 345]{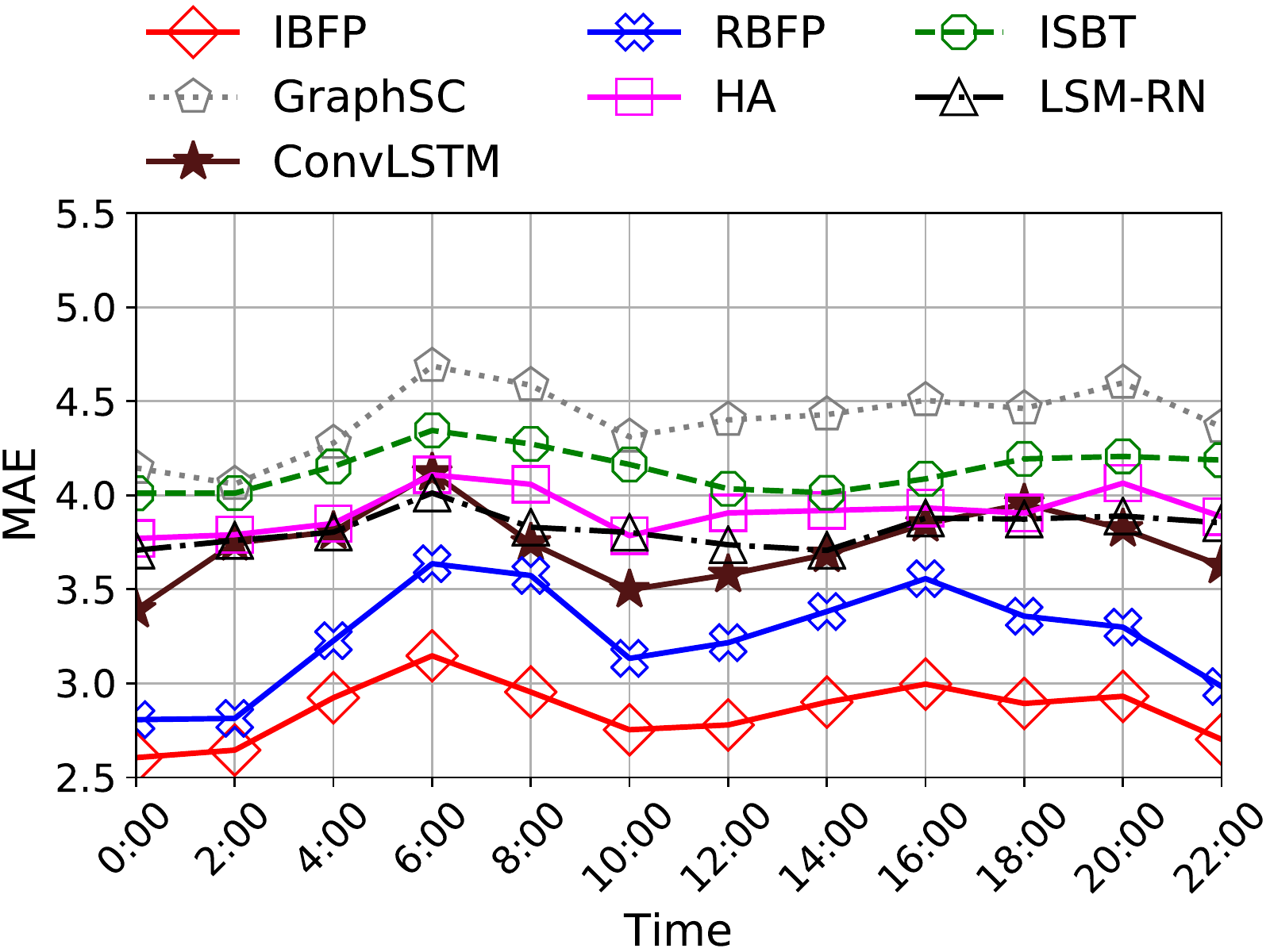}}
	\subfigure{
		\includegraphics[width=0.23\textwidth, bb= 0 0 459 345]{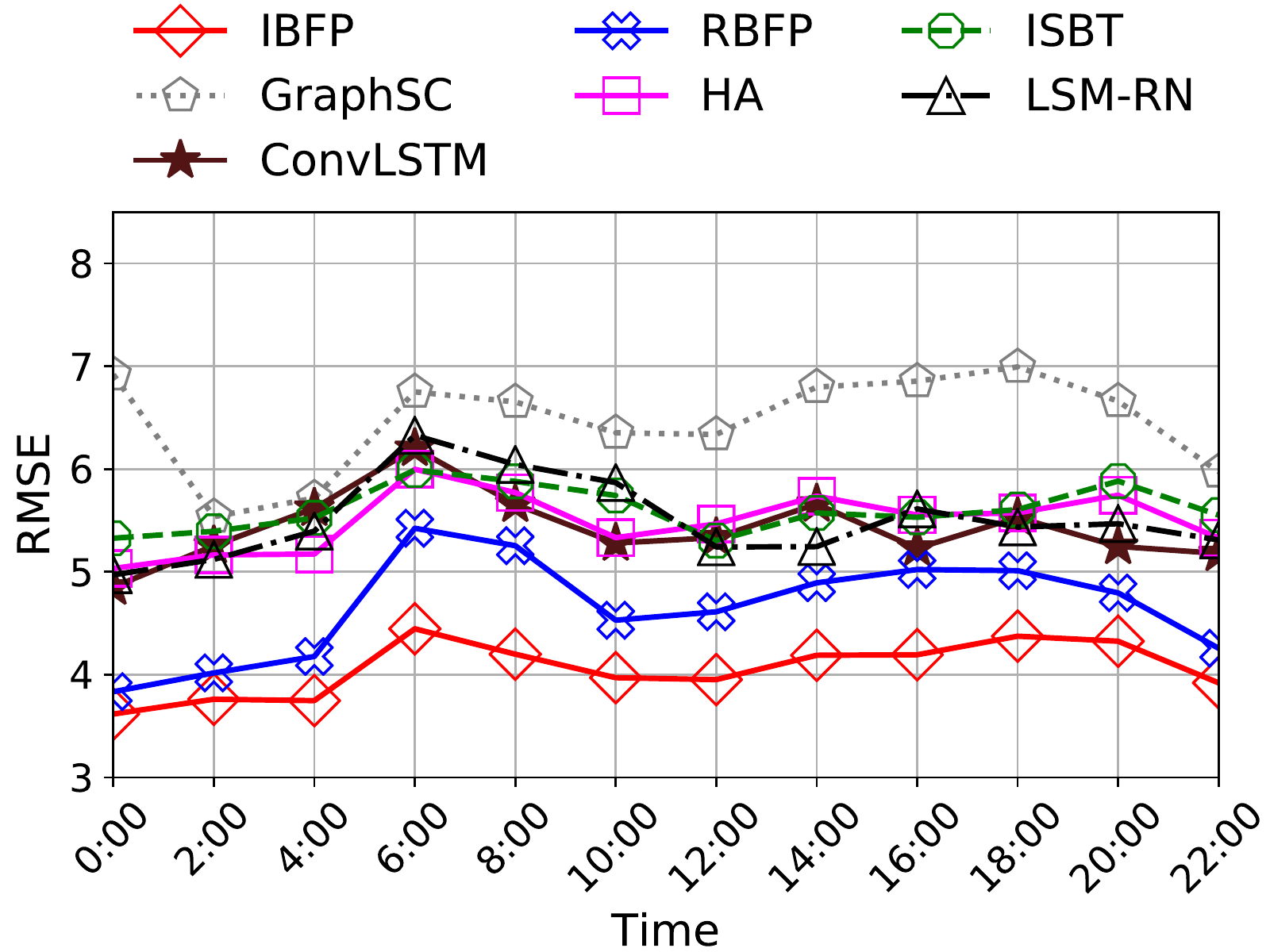}}
	\vspace{-3mm}
	\caption{Prediction errors on rainy working days}
	\vspace{-3mm}
    \label{res_dataset2}
\end{figure}

\begin{figure}[t]
	\vspace{-2mm}
	\centering
	\subfigure{
		\includegraphics[width=0.23\textwidth, bb= 0 0 459 345]{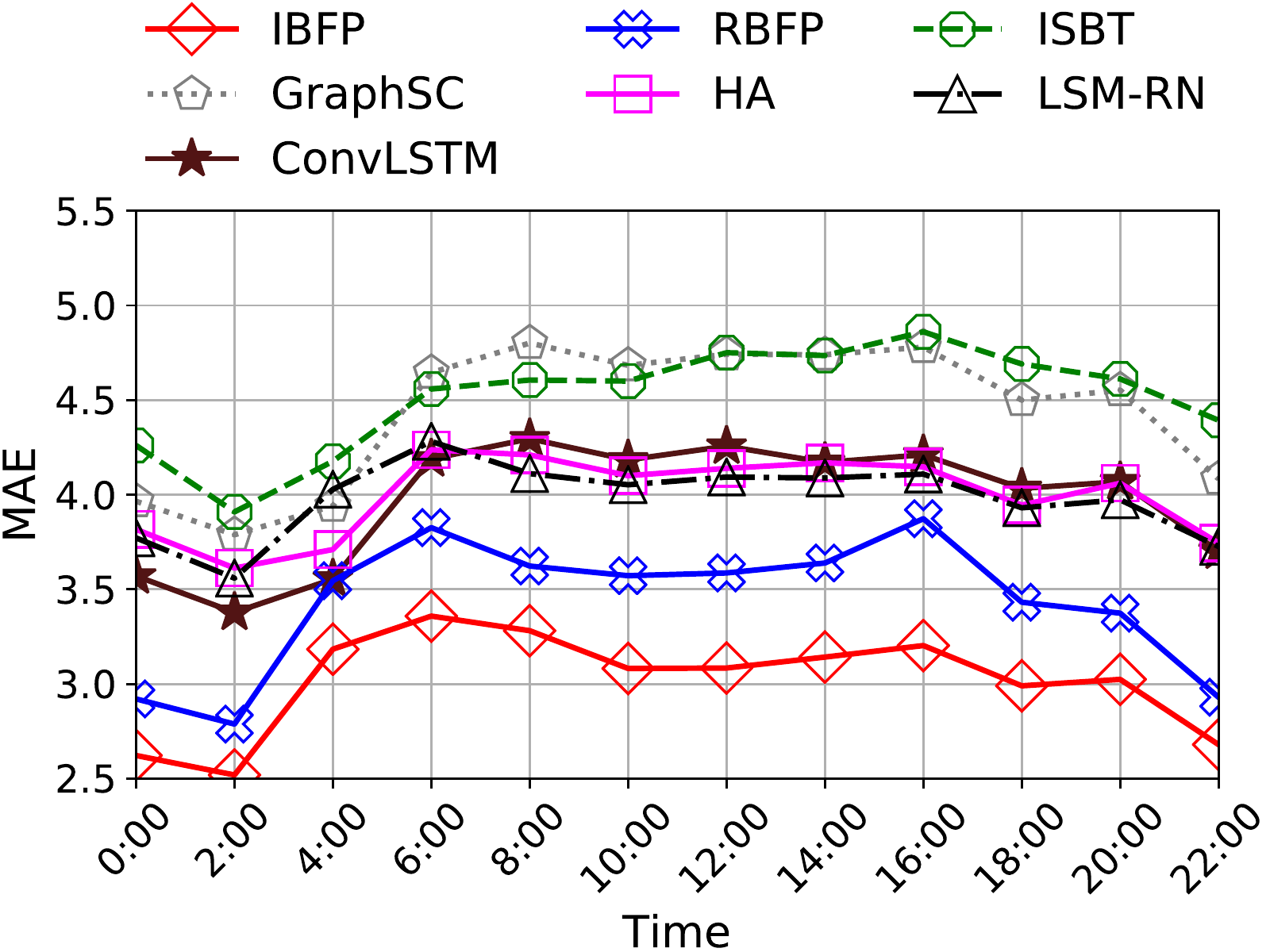}}
	\subfigure{
		\includegraphics[width=0.23\textwidth, bb= 0 0 459 345]{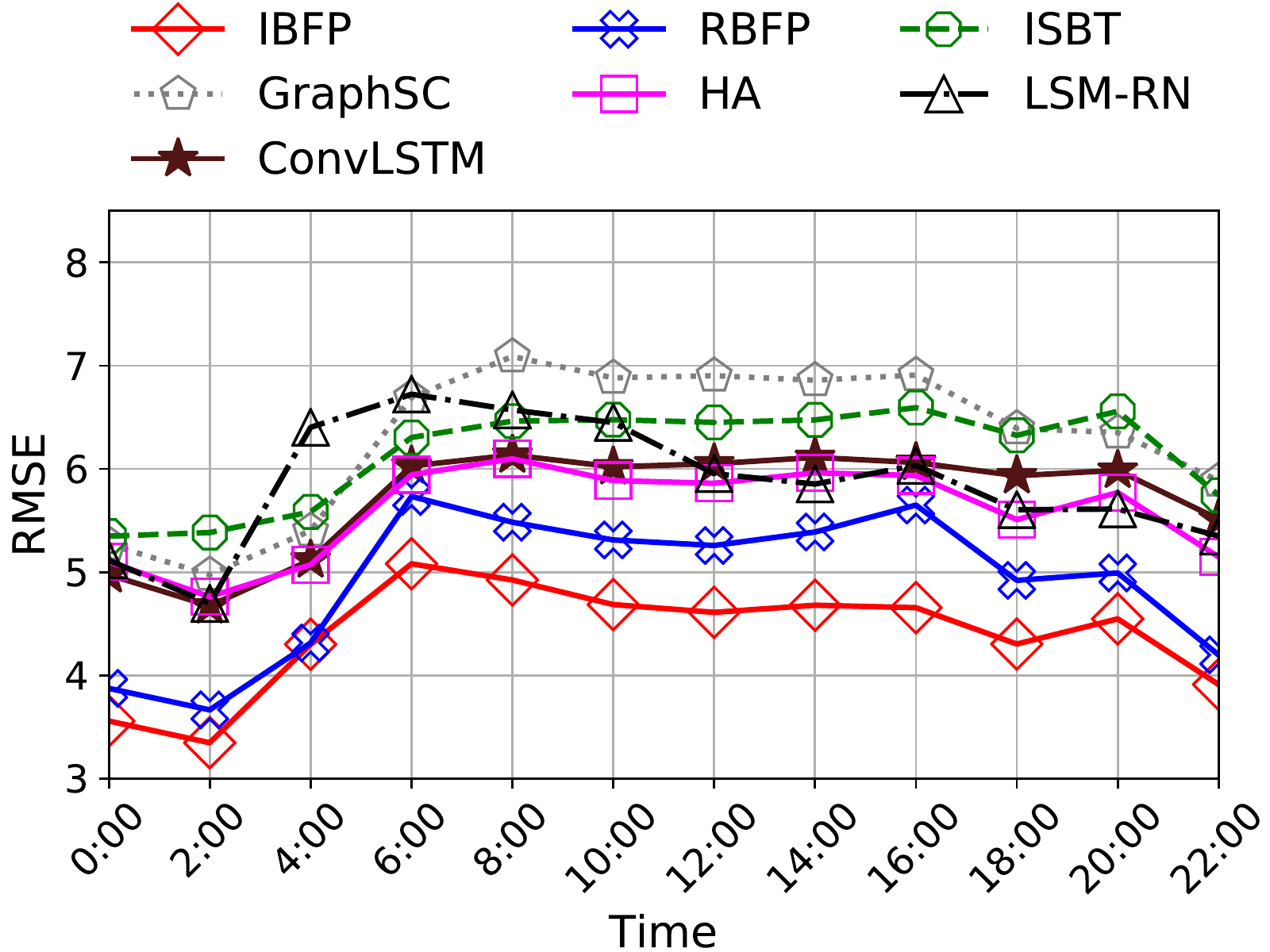}}
	\vspace{-3mm}
	\caption{Prediction errors on holidays}
	\vspace{-3mm}
    \label{res_dataset3}
\end{figure}

The overall performances of \graphsc are not satisfactory, since this model is not designed for dockless bikes, so does \lsmrn. \convlstm basically employs convolution operator to capture the spatial information at input-to-state and state-to-state transitions. However, convolution operator encodes local adjacent information, which is not very suitable in our scenario. For example, in the introduction, figure 1 shows zones near different exits of the same subway station (9 and 12), where the redundancy ratios are extremely imbalanced. Thus, the advantage of \convlstm could not be fully explored in our scenario. In addition, the training of Conv-LSTM is too costly, which limits the application scenario of this method.

\begin{figure}[t]
	\vspace{-2mm}
	\centering
	\subfigure{
		\includegraphics[width=0.45\linewidth, bb=0 0 1920 912]{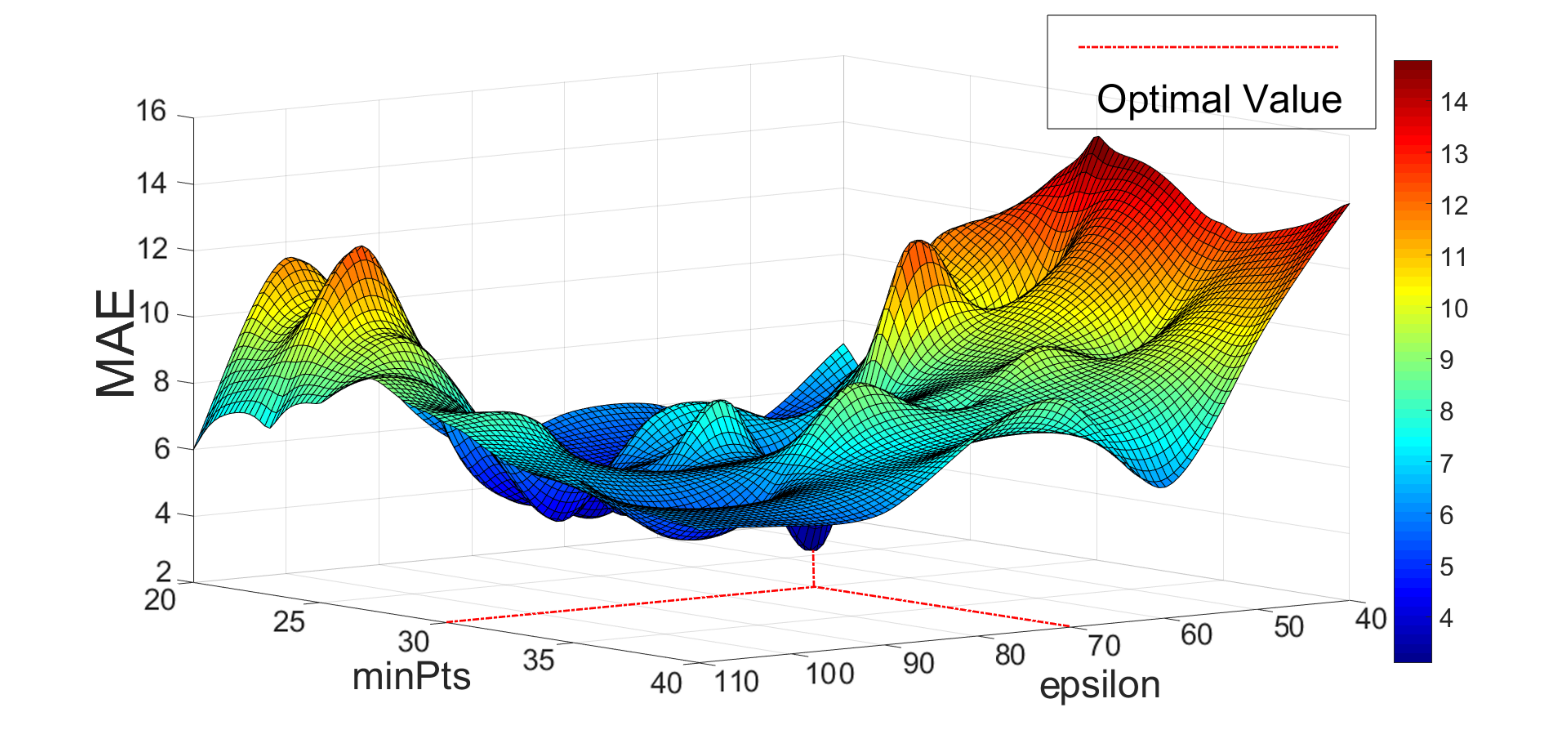}}
	\subfigure{
		\includegraphics[width=0.45\linewidth, bb=0 0 1920 912]{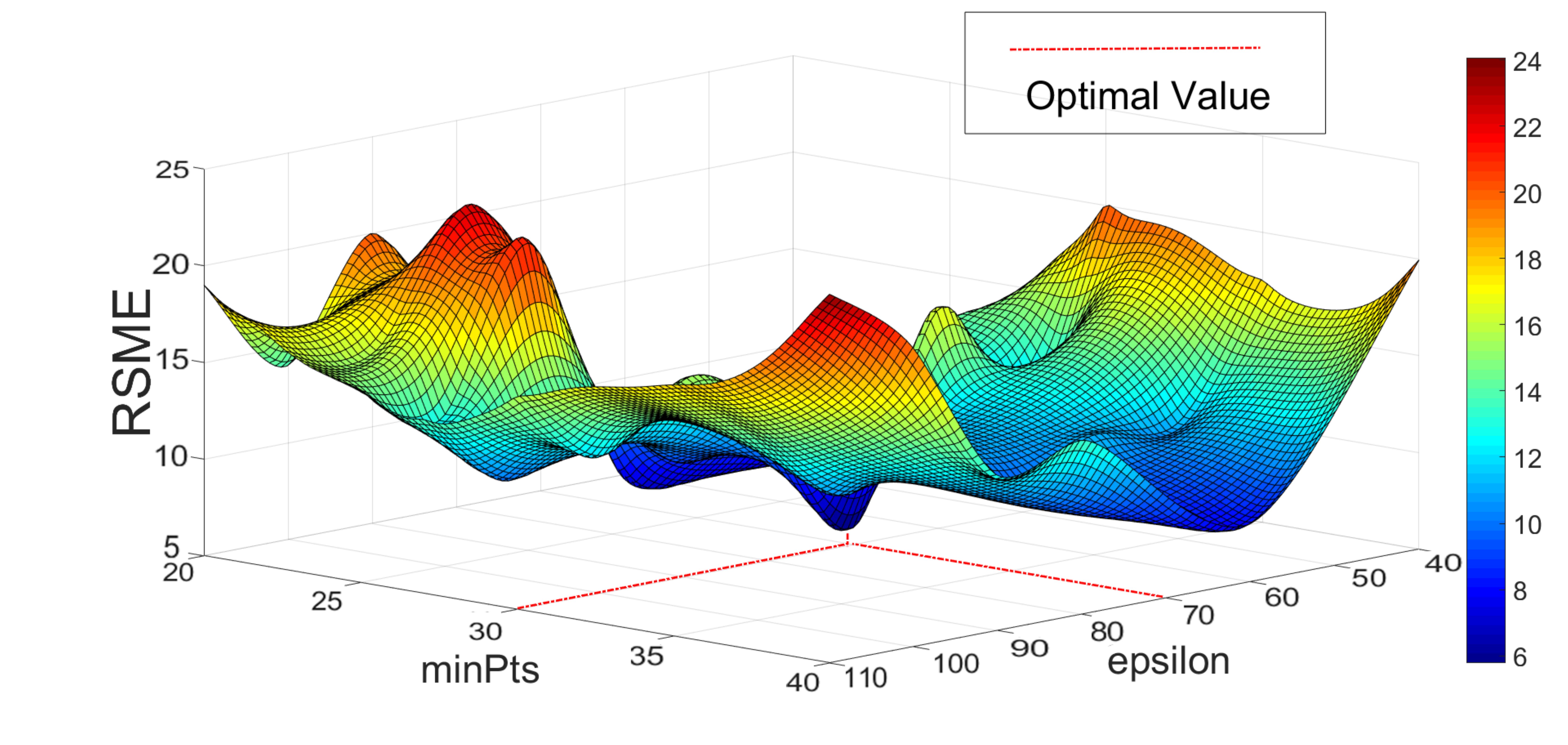}}
	\vspace{-3mm}
	\caption{Impacts of varying parameters $\epsilon$ and $minPts$ }\label{para_clustering}
	\vspace{-4mm}
\end{figure}

\begin{figure}[t]
	\vspace{-2mm}
	\centering
	\subfigure{
		\includegraphics[width=0.23\textwidth, bb=0 0 461 346]{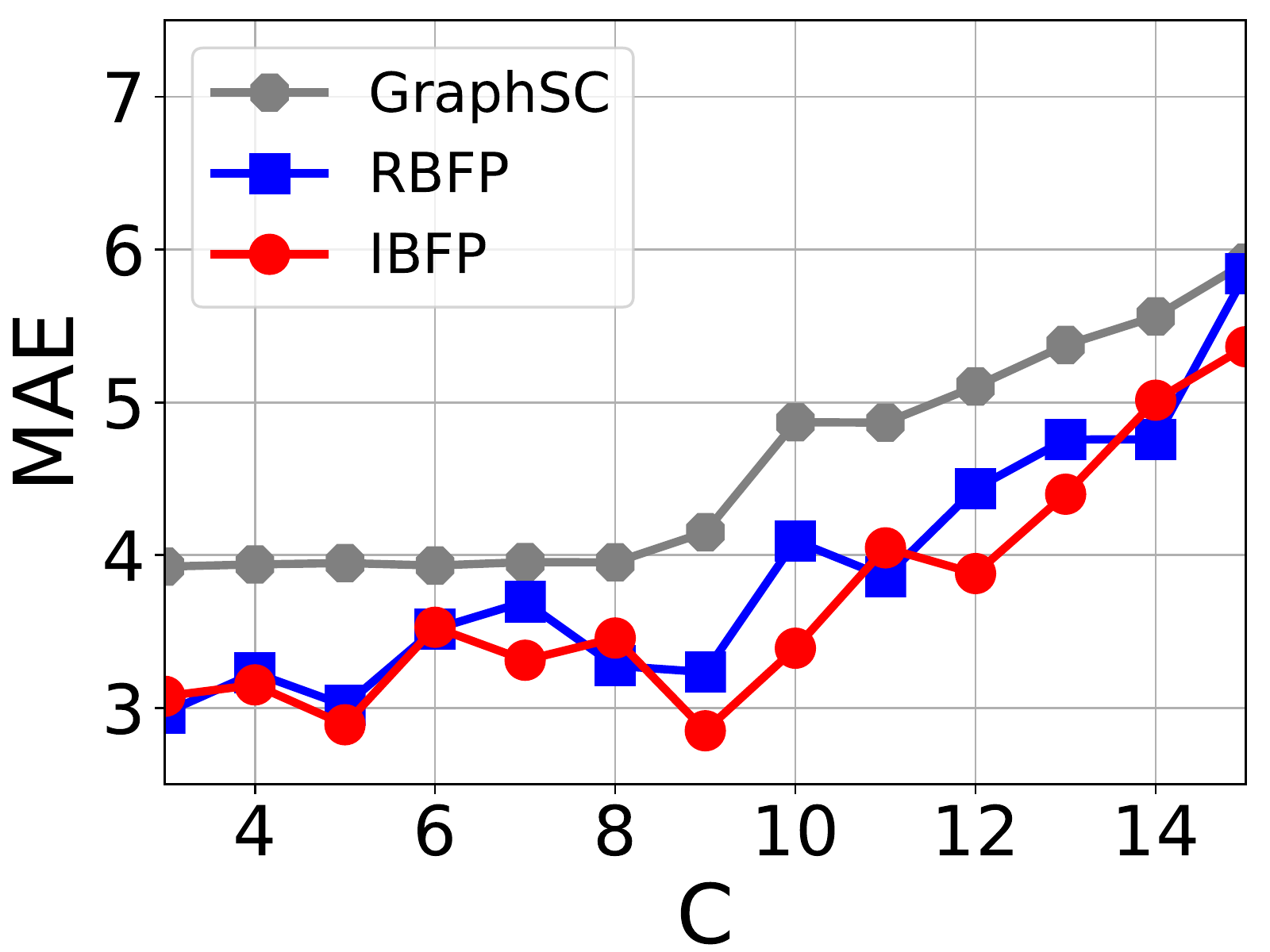}}
	\subfigure{
		\includegraphics[width=0.23\textwidth, bb=0 0 461 346]{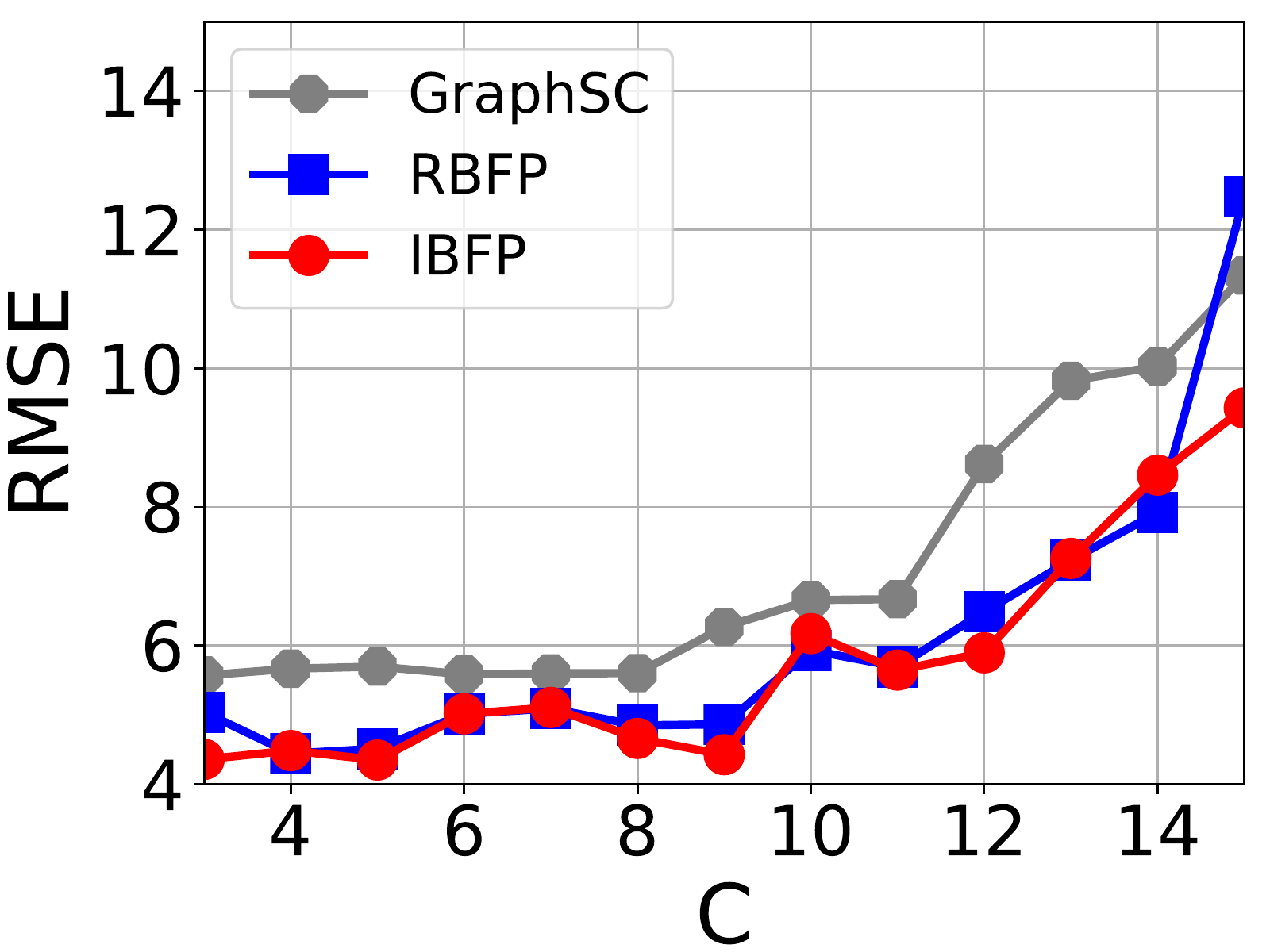}}
	
	\vspace{-4mm}
	
	\caption{Impacts of varying $C$ of basic matrices, regularization parameters $\lambda$ and $\gamma$ on prediction errors} \label{paras}
	\vspace{-2mm}
\end{figure}

\begin{figure}[t]
	\vspace{-2mm}
	\centering
	\subfigure{
		\includegraphics[width=0.23\textwidth, bb=0 0 720 540]{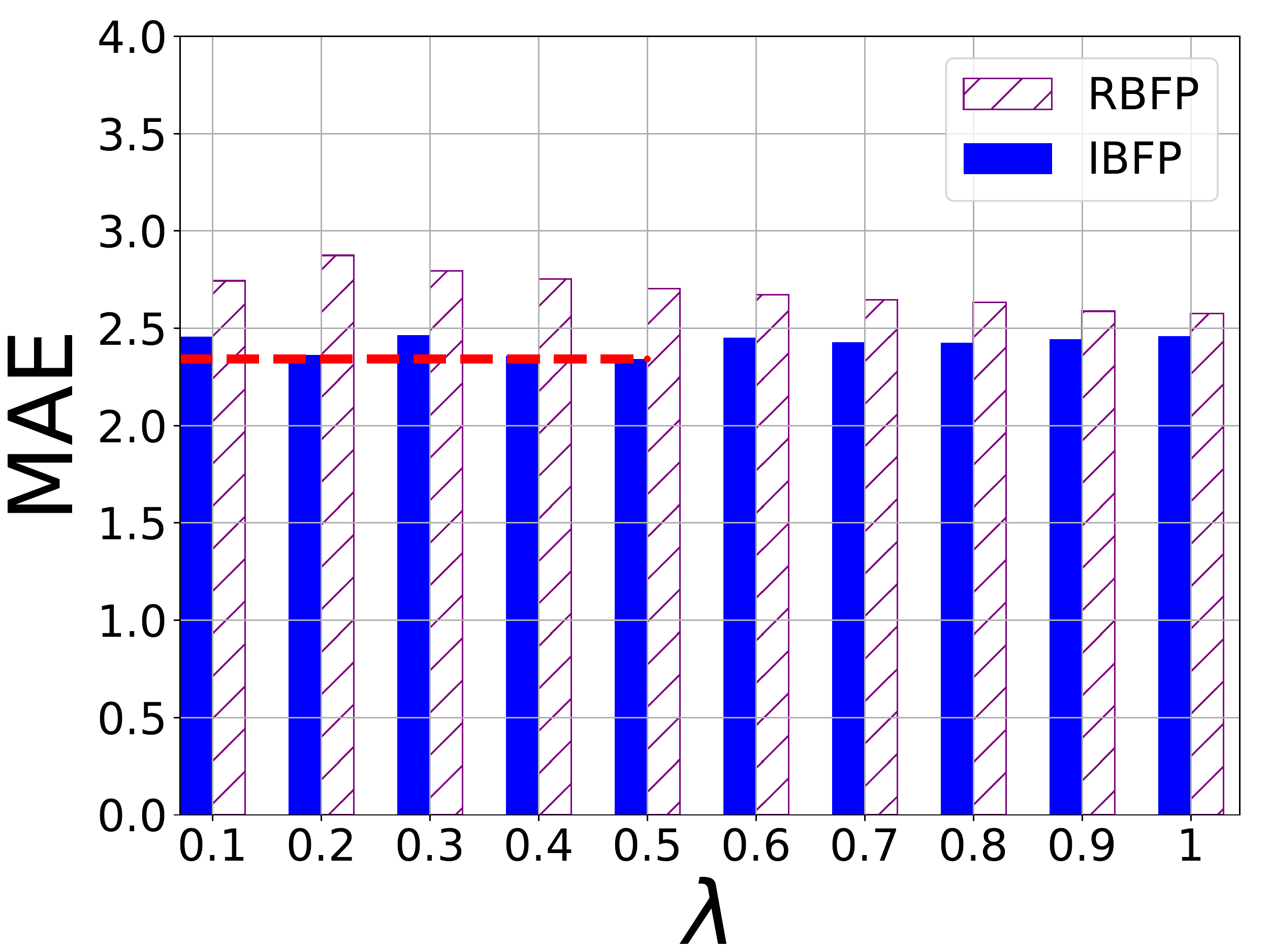}}
	\subfigure{
		\includegraphics[width=0.23\textwidth, bb=0 0 720 540]{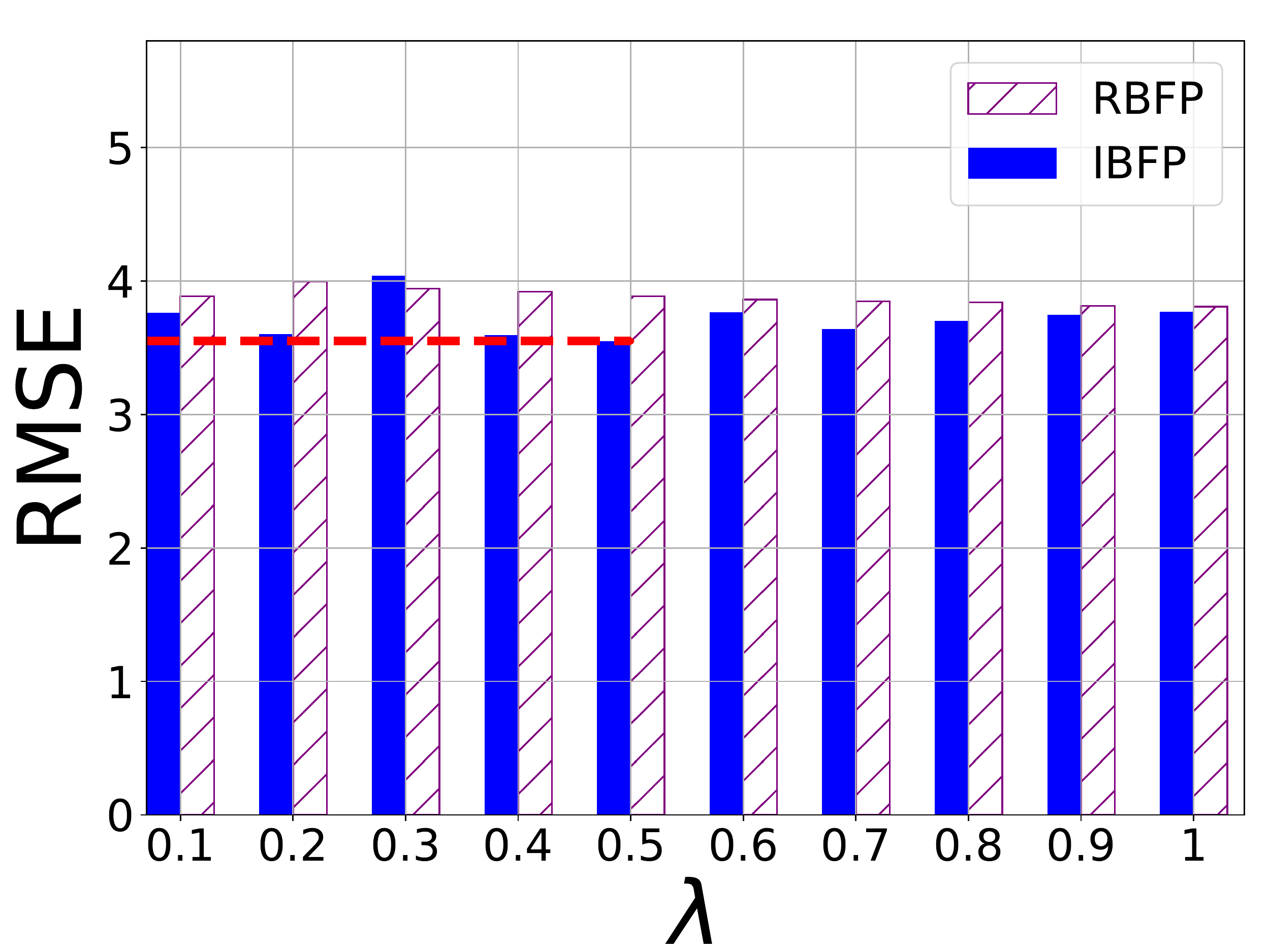}} \\
	\vspace{-4mm}
	\subfigure{
		\includegraphics[width=0.23\textwidth, bb=0 0 720 540]{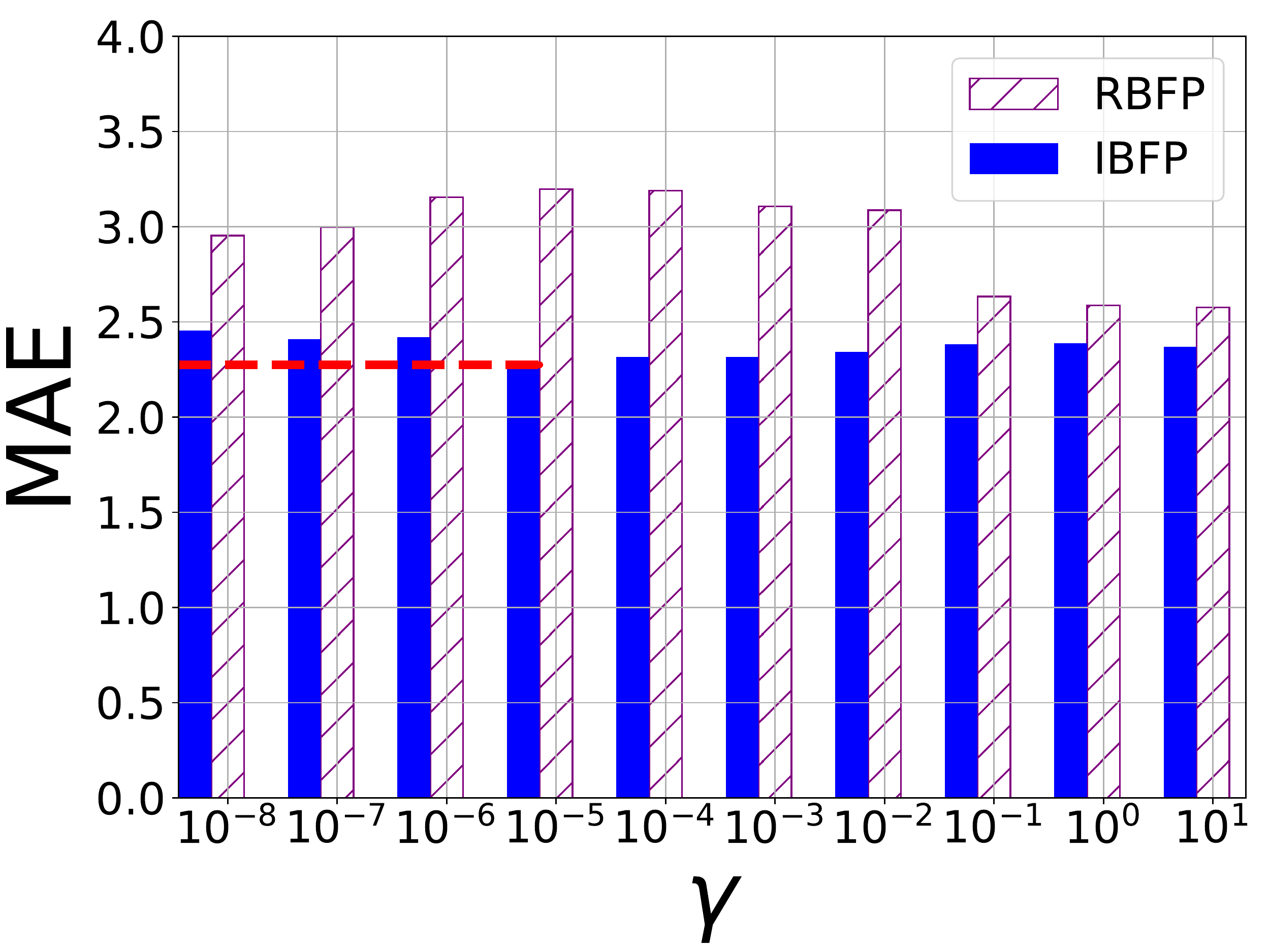}}
	\subfigure{
		\includegraphics[width=0.23\textwidth, bb=0 0 720 540]{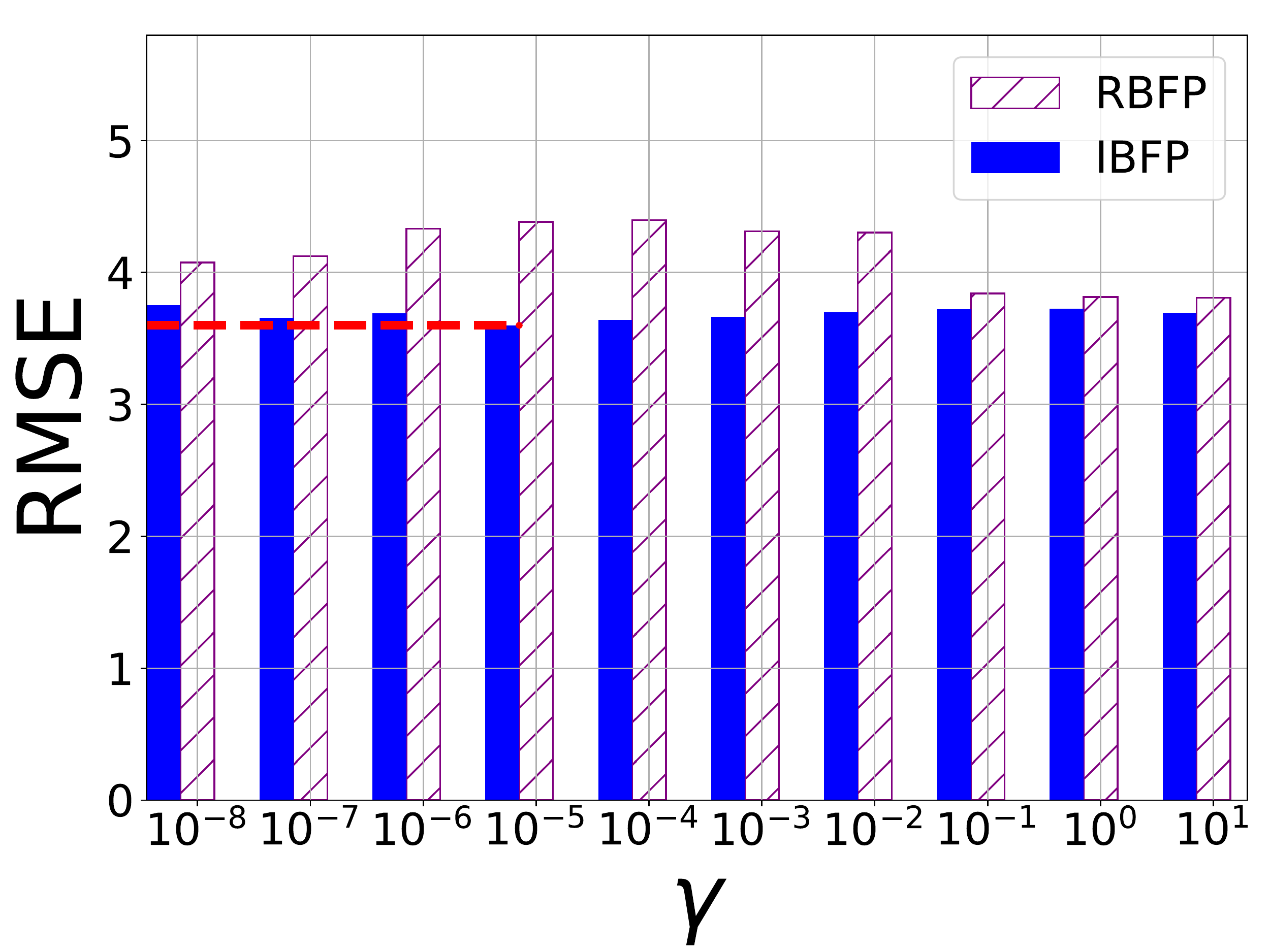}}
	\vspace{-3mm}
	\caption{Impacts of varying regularization parameters $\lambda$ and $\gamma$ on prediction errors} \label{para_Lamda}
	\vspace{-6mm}
\end{figure}

 \textbf{(2) Effectiveness on rainy working days.} 
We  evaluate the effectiveness on rainy working days when bike usage demands are concentrated on morning and evening peaks. Still, same as the data in working days, prediction results at around 6 a.m. are not also good enough due to the relocation of bikes. Again, all models are trained on the working day training set with parameters set to their default values. The prediction errors at different day time are reported in the \autoref{res_dataset2}.
For all algorithms, the general trends of prediction errors are similar to their counterparts on regular workdays. According to our results, the riding behavior in normal raining day is rarely affected. This is somehow counter-instinct as we notice that the heavy raining surely changed the riding behavior. The explanation is in the testing data there is few data connect to such weather, while the effect of normal or shot time raining is not explicit. However, as indicated by a few increased errors, the prediction on rainy days is still non-trivial, possibly because there are more factors involved in bike traffic flows on rainy days. Despite the difficulty, our \ourmodel model still consistently performs the best in all our tests. 

 \textbf{(3) Effectiveness on holidays.} 
We further evaluate the effectiveness on holidays when bike usage demands are more dispersed than on working days and people are generally willing to ride bikes. All models are trained on the holiday training set with parameters set to their default values. The prediction errors at different day time are reported in \autoref{res_dataset3}.

The general trends of prediction errors on holidays are slightly different from those on working days in the sense that the errors have an obvious drop between the morning and evening peaks on working days while they remain in a relative high level between the two peaks on holidays. As we state before,  bike usage demands on holidays are more dispersed and random, which makes the prediction a bit more challenging. This also justifies our separation of working day and holiday data that govern different patterns. Again, our \ourmodel model consistently performs the best in all our tests. 

From these figures, we also can observe that the errors of our multi-step prediction results slowly increase as the number of prediction steps increases, especially in the day time (8:00 - 20:00) in these figures. However, at the first predicted time ( 00:00 in the midnight ) and the last predicted time (22:00 in the night), the prediction error recovered to some extent due to very few riding features.


\vspace{-4mm}

\subsection{Discussion} \label{exp-discussion}

\textbf{(1) Analysis of the baselines.} 
Here we expand significantly in detailed description of why some baselines are not as good as ours in these scenarios.

\textbf{ISBT}, first of all, is mainly for the rebalancing problem of dock-based shared bikes, and we just chose the part of flow prediction. 
Second, ISBT uses a Gaussian mixture model to fit the distribution of inter-regional riding distances from the sharing bike data of New York City. By comparing the NYC data with our data in Shanghai, we found that the average riding distance or time of NYC's was significantly longer than ours. Statistics of bike trips of our data had been presented in the Figure 5, whose travel time and distance are too short (usually ten or twenty minutes) to fit the appropriate travel time model  well. In summary, the advantages of ISBT are not obvious when used for our data and scenario.

\textbf{ConvLSTM}, is a typical prediction model based on Convolutional Neural Network (CNN) by spatio-temporal characteristics. However, our work focused on the flow between region-region pairs. In the experiments, we basically partitioned 768 square regions for the ConvLSTM, which implies that there are 768 channels in the training simultaneously. Thus, although ConvLSTM has a good performance on prediction, it is too complex in training phase to get accurate results in our scenario, as it requires independent training for each pair of regions. The increase of the number of channels will cause the model complexity souring rapidly. With dozens of experimental attempts of ConvLSTM, we found the gradients vanished to a large extent in the training phase and more than 6 channels will make the experiment intractable.

\textbf{LSM-RN}, also establishes a matrix res-construction problem. However, since it predicts the speed of the traffic through perceived data of sensors, its time interval is set to 5 minutes (our time interval is 1 hour). So it only uses one identical transfer matrix (similar to the matrix A in our paper) for every 20 consecutive time segments. Obviously, for our task, data in 20 consecutive 1-hour segments contains different traffic patterns, such as morning and evening peaks, off-peak, and late night, which can not be represented by only one transfer matrix. Furthermore, it does not exploit such traffic features as we do.

\textbf{GraphSC}, also learns the graph regularized sparse representation. However, GraphSC is originally designed for the image representation and utilizes the SVM to perform the classification prediction which is too far from our problem. Thus, although with great efforts for improvement, the performance of this method is far worse than our method in the scenario of our work.

\textbf{(2) Inapplicability.} 
Some cases where our methods may not be applicable: bad weather, especially the heavy rain, or rainstorm. This is mainly because that the city we studied, Shanghai, only has few days of this extreme weather in the whole year. In fact, light-rain days are the most rainy days in our data, whose impact on the bike flow is not serious. Therefore, no matter how to generate the training and testing sample sets, there is always not enough data to learn meaningful patterns in extreme weather. Furthermore, our approach does not respond enough well to changes in traffic due to some unusual or unexpected situations, especially in high-flow areas in high-flow areas (Detailed explanation will be represented in sub-setion 4.5.1 ).

\vspace{-1mm}
\subsection{Parameters Sensitivity} \label{exp-para}

In the second set of experiments, we test the sensitivity results on (1) parameters $epsilon$ and $minPts$ of DBSCAN clustering used in region segmentation (2) the number $C$ of base matrices, and (3) the regularization parameters $\lambda$ and $\gamma$ of the loss function. We perform the cross-validation for training these parameters. 
We vary different values of parameters, and fix the rest parameters to their default values: $C$ of base matrices is set to 9, $epsilon$ and $minPts$ are fixed to 70m and 30, regularization parameters $\lambda$ and $\gamma$ are set to 0.4 and $10^{-5}$, $p$ and $q$ in Section~\ref{prediction} are set to 5 and 8.

 \textbf{(1) Impacts of} $\epsilon \& minPts$.
Recall that our model first divides the entire city into regions with a density-based clustering method, \ie DBSCAN. 
A key concept of the DBSCAN algorithm is core points which have at least $minPts$ points within distance $epsilon$.  
To evaluate the impacts of $epsilon$ and $minPts$ \cite{erman2006traffic}, we vary $epsilon$ from 40m to 110m and $minPts$ from 20 to 40. 
The results are reported in the \autoref{para_clustering}. 
When $epsilon$ and $minPts$ varies, the prediction errors also change under our two metrics, which indicates a significant impact of the divided regions on bike flow prediction. Both $epsilon$ and $minPts$ influence the prediction accuracy in a complex manner. However, the best accuracy is obtained under the same setting, \ie  $epsilon=70m$ and $minPts=40$, with (\mae and \rmse) being (1.8172 and 5.1217), respectively.

 \textbf{(2) Impacts of $C$.}
To evaluate the impacts of the number $C$ of base matrices, we vary $C$ from 3 to 15, 
and test the prediction errors of the three algorithms that utilize sparse representation, \ie \graphsc, \stbfp and \ourmodel. The results are reported in the \autoref{paras}. When $C$ varies, the prediction errors of \graphsc keep increasing while the ones of \stbfp and \ourmodel first fluctuate and then increase. Moreover, our \ourmodel model  is consistently better than \graphsc when using the same number of base matrices, and is generally better than \stbfp. Similar to varying $epsilon$ and $minPts$, the best accuracy of \ourmodel is obtained when fixing $C=9$, with (\mae and \rmse) being (2.8502 and 4.4264), respectively. These are indeed the best accuracy that can be obtained by all of the three models.

 \textbf{(3) Impacts of regularization parameters $\lambda$ and $\gamma$.}
Recall in \autoref{eqn:06} that parameters $\lambda$ and $\gamma$ regularize the strength of the sparsity constraint and the temporal constraint, respectively. 
To evaluate the impacts of $\lambda$, we vary $\lambda$ from 0.1 to 1.0, 
and test the errors of our model \ourmodel and its variant \stbfp. We do the same for parameter $\lambda$ except that we vary $\lambda$ from $10^{-8}$ to 10. The results are reported in \autoref{paras}.
When varying $\lambda$ and $\gamma$, the prediction errors of both \stbfp and \ourmodel vary in a narrow range, our \ourmodel model is consistently better than its simplified version \stbfp in all our tests. Again, the best accuracy of \ourmodel is obtained when $\lambda=0.5$ and $\gamma=10^{-5}$, no matter which metric is used.

\vspace{-2mm}

\subsection{Case Studies}

\iftrue

\subsubsection{Prediction Accuracy}
\definecolor{royalblue}{rgb}{0.05,0.6,1} 
\begin{figure}[t]
	\vspace{-4mm}
	\ \ \ \ \ \ \ \ \small{\sf{\tikz\draw[red,fill=red] (0,0) circle (.5ex);  high errors \hspace{3ex}  \tikz\draw[yellow,fill=yellow] (0,0) circle (.5ex); moderate errors \hspace{3ex}  \tikz\draw[royalblue,fill=royalblue] (0,0) circle (.5ex); low errors }} \\
	\subfigure[working day]{
		\includegraphics[width=0.23\textwidth, bb=0 0 596 842]{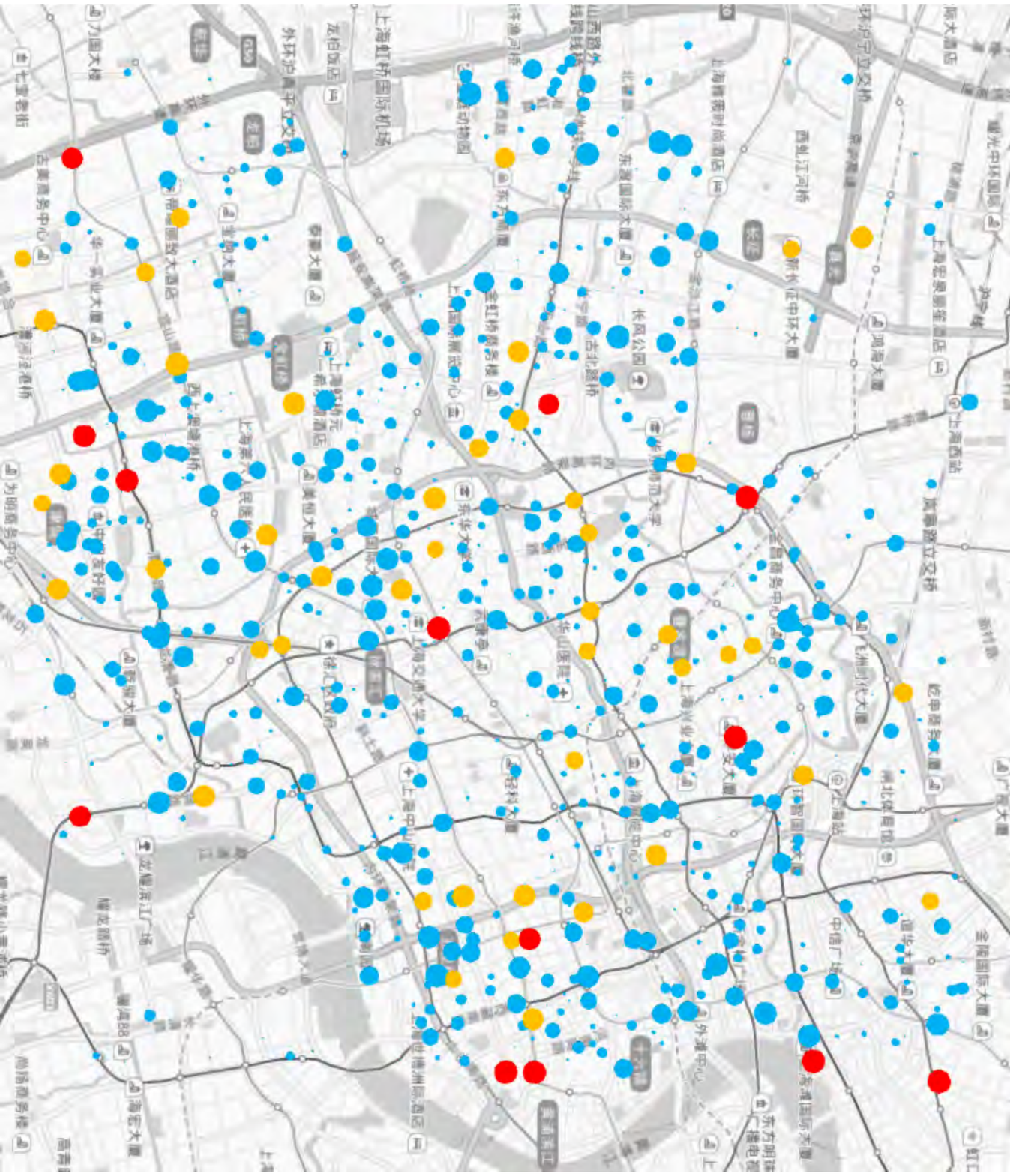}} 
     \subfigure[holiday]{
        \includegraphics[width=0.23\textwidth, bb=0 0 596 842]{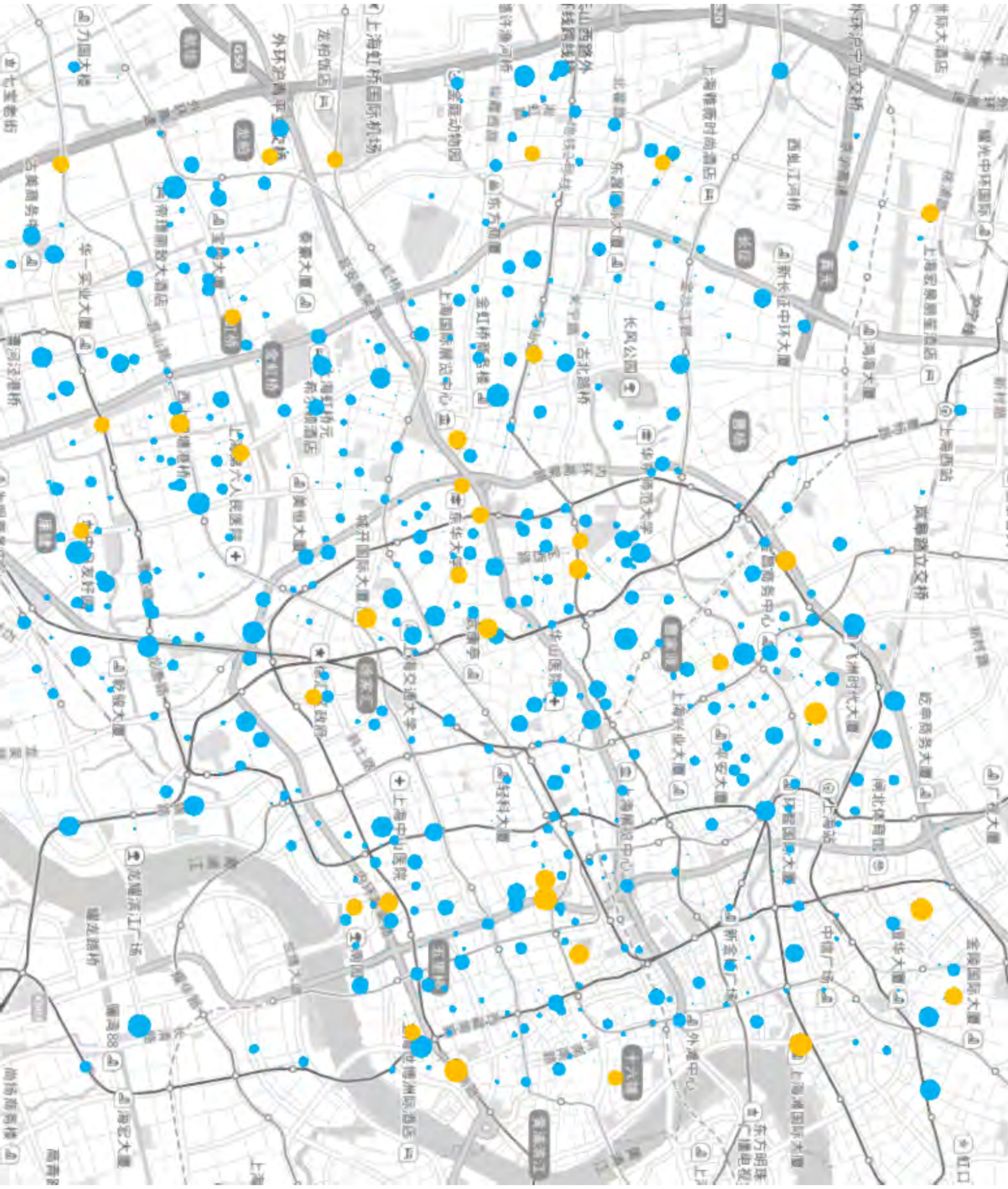}}
	\vspace{-2mm}
  	\caption{An example of the prediction errors.}
	\vspace{-6mm}
    \label{case:error}
\end{figure}

\autoref{case:error} illustrates the prediction errors of pick-up flows during 8:00 a.m. and 9:00 a.m. on a randomly selected working day and holiday in Shanghai, respectively. For presentation purpose, we compute the pick-up flow at each region during the concerned period of time, \ie the sum of each row in the flow matrix. For both figures, different colors indicate different error levels, and a larger filled circle further indicates a higher error within the corresponding level. 
As can be seen from the figures, the circles in the high errors and low errors are quite similar in terms of locations on both the working day and the holiday. 
By matching these regions on the map of Shanghai, we find that: on the working day, high flow regions mainly locate around hospitals, CBDs and governmental buildings; on the holiday, the number of regions having high flows significantly decreases and the remaining regions mainly locate around large hospitals such as Huashan hospital and Shanghai Sixth People's Hospital.
This information is very useful for rebalancing bikes and, hence, running an effective bike share systems. Specifically, from the figure, we find some error-prone areas, which are mainly high-flow areas in high-flow areas, especially with entertainment and leisure as the main area functionalities, such as People's Square and Nanjing Road Subway Station. We consider that bike flows in these areas are vulnerable to some additional factors, such as irregular commercial activities and traffic accidents. That is, our approach does not respond enough well to changes in traffic due to these unusual situations.

\fi
\vspace{-3mm}
\subsubsection{Interperablility of Base Matrices.}
\begin{figure}[t]
\setlength{\abovedisplayskip}{2pt}
\setlength{\belowdisplayskip}{0pt}
\centering

		\includegraphics[width=0.41\textwidth, bb=0 0 834 392]{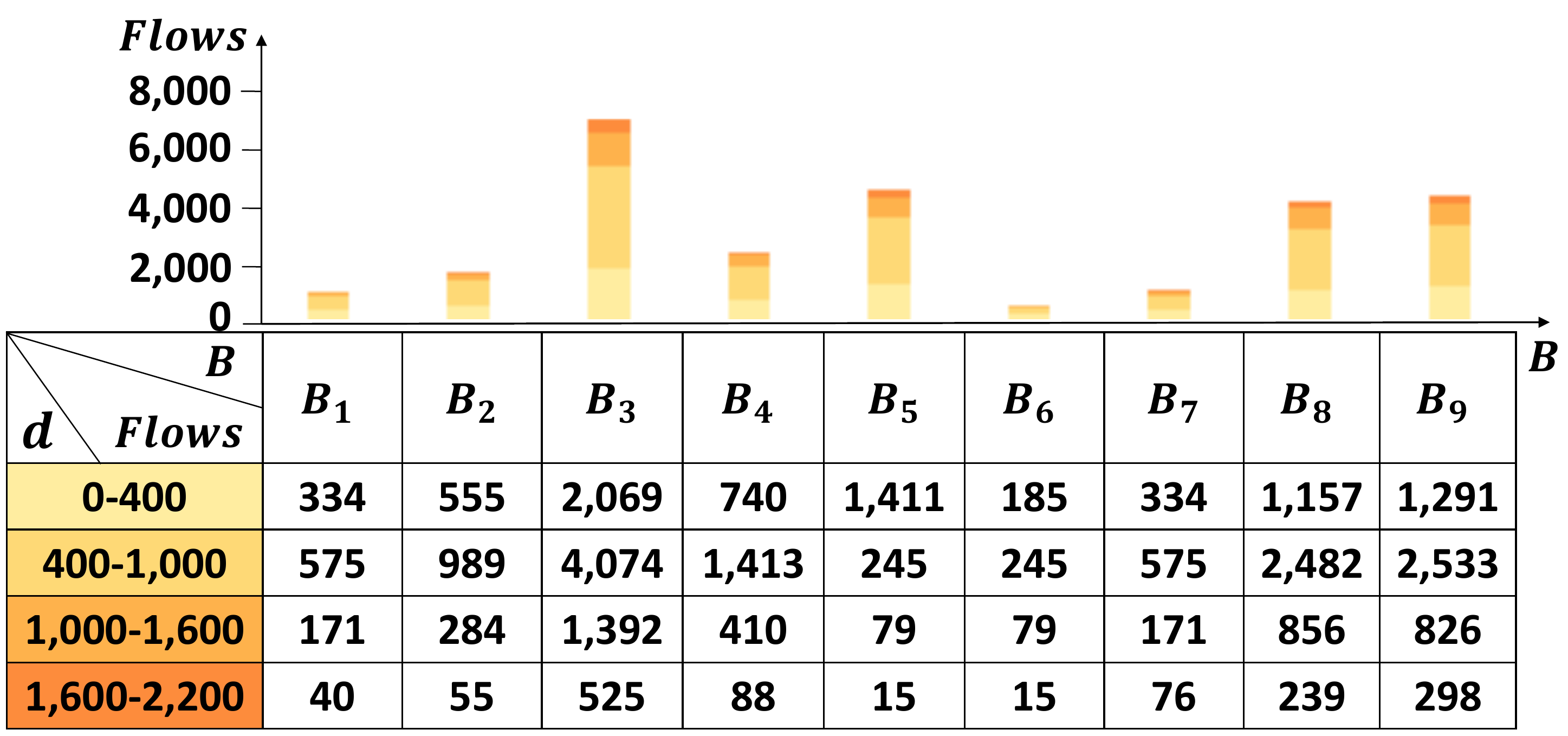}
\vspace{-2mm}
\caption{Frequency statics of base matrices in different straight riding distances.}
\vspace{-4mm}
\label{case:base}
\end{figure}

\begin{figure*}[t]

\vspace{-2mm}

\centering
	\subfigure[Working days rush hours.]{
		\includegraphics[height=1.8 in, width=3.5 in, bb=0 0 653 358]{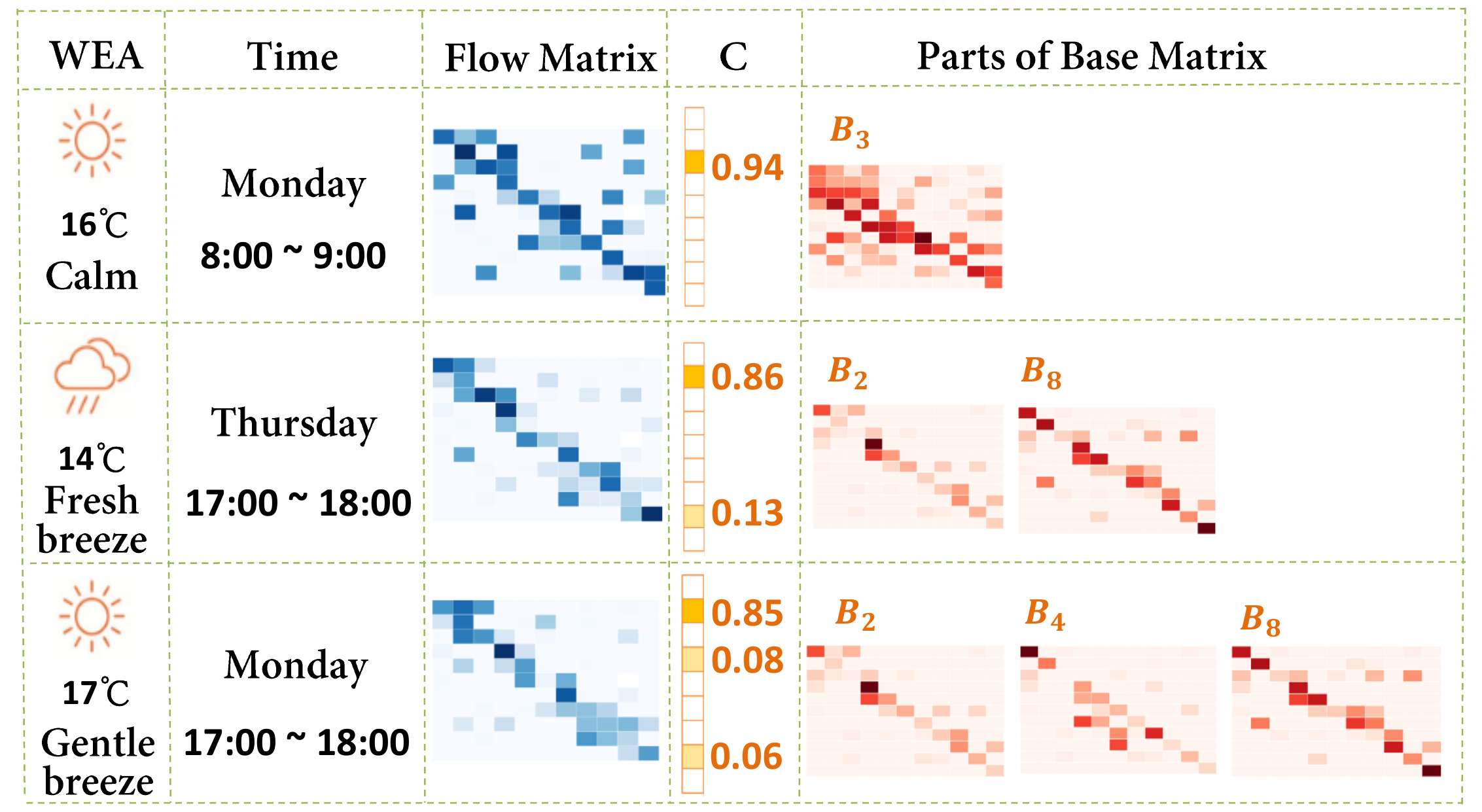}} 
		\hspace{-2ex}
	\subfigure[Weekend nights.]{
		\includegraphics[height=1.8 in, width=3.5 in, bb=0 0 653 264]{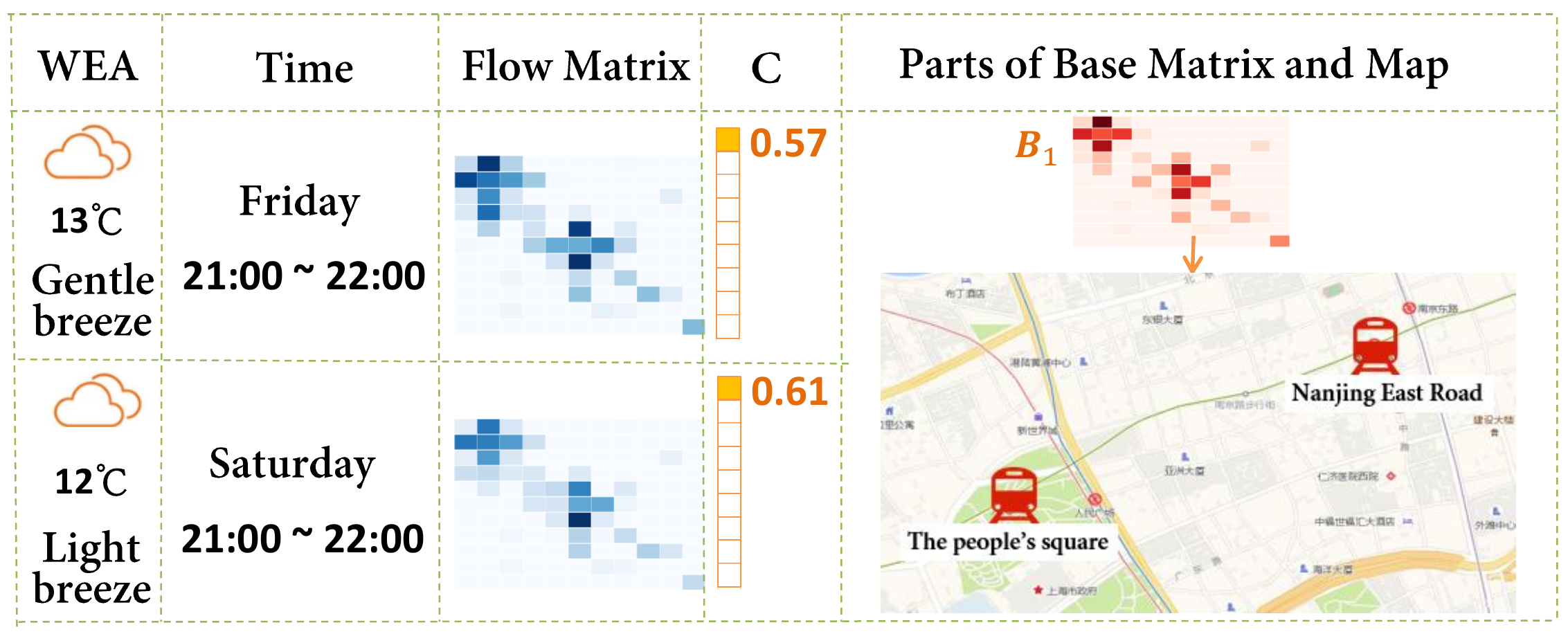}} 
	\vspace{-2mm}
\caption{Examples of flow matrix reconstruction with base matrices and its interpretability.}

\vspace{-3mm}

\label{case:interpretability}
\end{figure*}

To illustrate the interpretability of our method, we give a case study here.  \autoref{case:base} shows the frequency statics of all base matrices in different straight riding distances ($d$ meters in the figure), where, the histogram visually presents the total bike flow and composition of the base matrices in distance interval. For example, $B_3$ and $B_6$ stand for the base matrix with maximum and minimum of the total flows. It can be observed that $B_3$ consists of maximum number in long riding distance intervals, which is marked in the darkest color in the figure. We further illustrate the interpretability of base matrices in \autoref{case:interpretability} with case studies, where we show how the base matrices compose traffic patterns in different environment settings. In \autoref{case:interpretability} (a), the first and second column stand for the weather and time of the third column's flow matrices, which could be approximately reconstructed with the fourth column's coefficient vectors ($C$) and the nine base matrices. Note that, in the fourth column, we only list part of the base matrices/matrix with large coefficients $[0.01, 1)$. In addition, because the size of a whole flow matrix (the same as a base matrix) is 766 $\times$ 766, it is hard to be clearly visualized. Here, we just present the distribution of a small part of the whole matrix.

Particularly, in \autoref{case:interpretability} (a), matrices represent the traffic flow in a downtown zone surrounding the Changshou road in Shanghai city, including residential area, office buildings, subway stations, malls and schools. The first flow matrix represents a regular morning peak (Monday, 8 a.m. to 9 a.m.) in a normal weather (sunny, 16 Celsius). Here, we can find the largest reconstruct coefficient $s_3= 0.94$, which indicating the rush hour pattern could be approximate with the base matrix $B_3$ only. In the other words, the base matrix $B_3$ could be interpreted as the rush hour base matrix.  
Comparing the second and third flow matrices, which are the late rush hours but in different weather environment(rainy and sunny), we find some interesting results: 1) the distribution of the original flow matrices are similar no matter in rainy or sunny weather. This is because it was not a heavy rain (just light rain) at this time and in this zone, the bike user's demand was very high. 2) The highest reconstruct coefficients are $s_2$, which denotes the second one of the coefficient vector. This coefficient associates with $B_2$, which is found to cluster most flow matrices of late rush hours, while following the same way, we find that $B_8$ could presents the flow pattern from the work region to the residential area. 3) Though the rainy day does not have much impact on overall bike flow in this case, it does influence the user's activity modes. Comparing the second and third flow matrices and the corresponding coefficients, we find that the value of $s_8$ and $s_4$ are the key difference. In our case, $B_4$ is representing the flow to the recreation area. Thus, possible explanation for the third flow matrix is that better weather, like sunny, leads to more recreation activities, e.g. dinner, fitness, movies and etc.

We can also find some interesting patterns regarding the weekend flows, some of which are illustrated in the \autoref{case:interpretability}(b). In this figure, both rows of matrices represent traffic pattern of the weekend nights, from which we can find that the majority component of them is $B_1$. Comparing with other areas, the majority traffic flows are around two hot spots, which are The People's Square and Nanjing East Road respectively. These two places contain many fashion malls, large subway stations and theaters. These findings reveal the common facts that in weekends, recreation activities are more centered and last late at night.

\vspace{-3mm}
\iftrue
\subsubsection{Analysis of User Mobility}

\begin{figure*}[t]
	\centering
	\subfigure[Pick-up, Working day]{
		\includegraphics[height=1.3in,width=1.4in, bb=0 0 810 800]{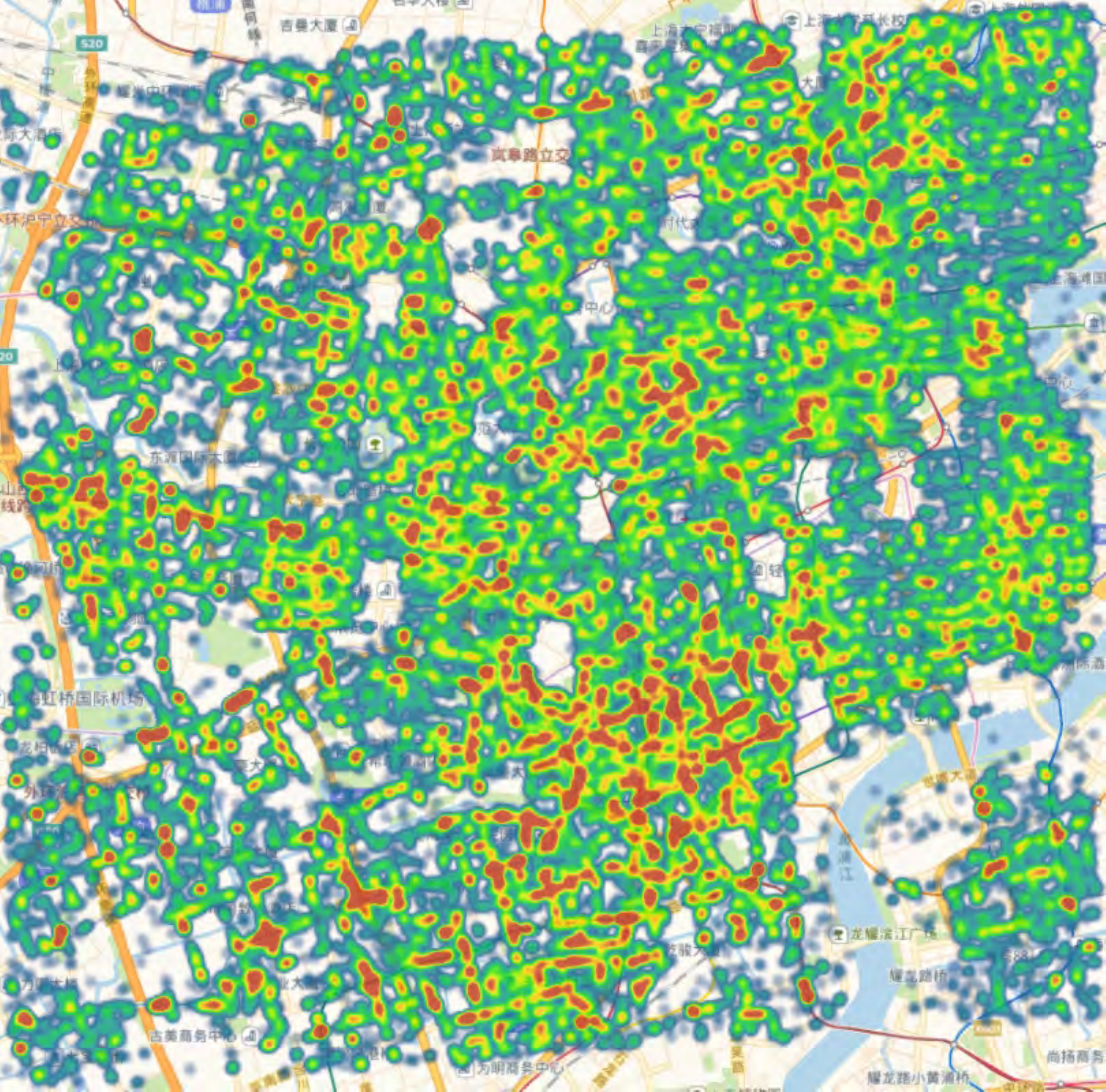}} 
	\subfigure[Drop-off, Working day]{
		\includegraphics[height=1.3in,width=1.4in, bb=0 0 810 800]{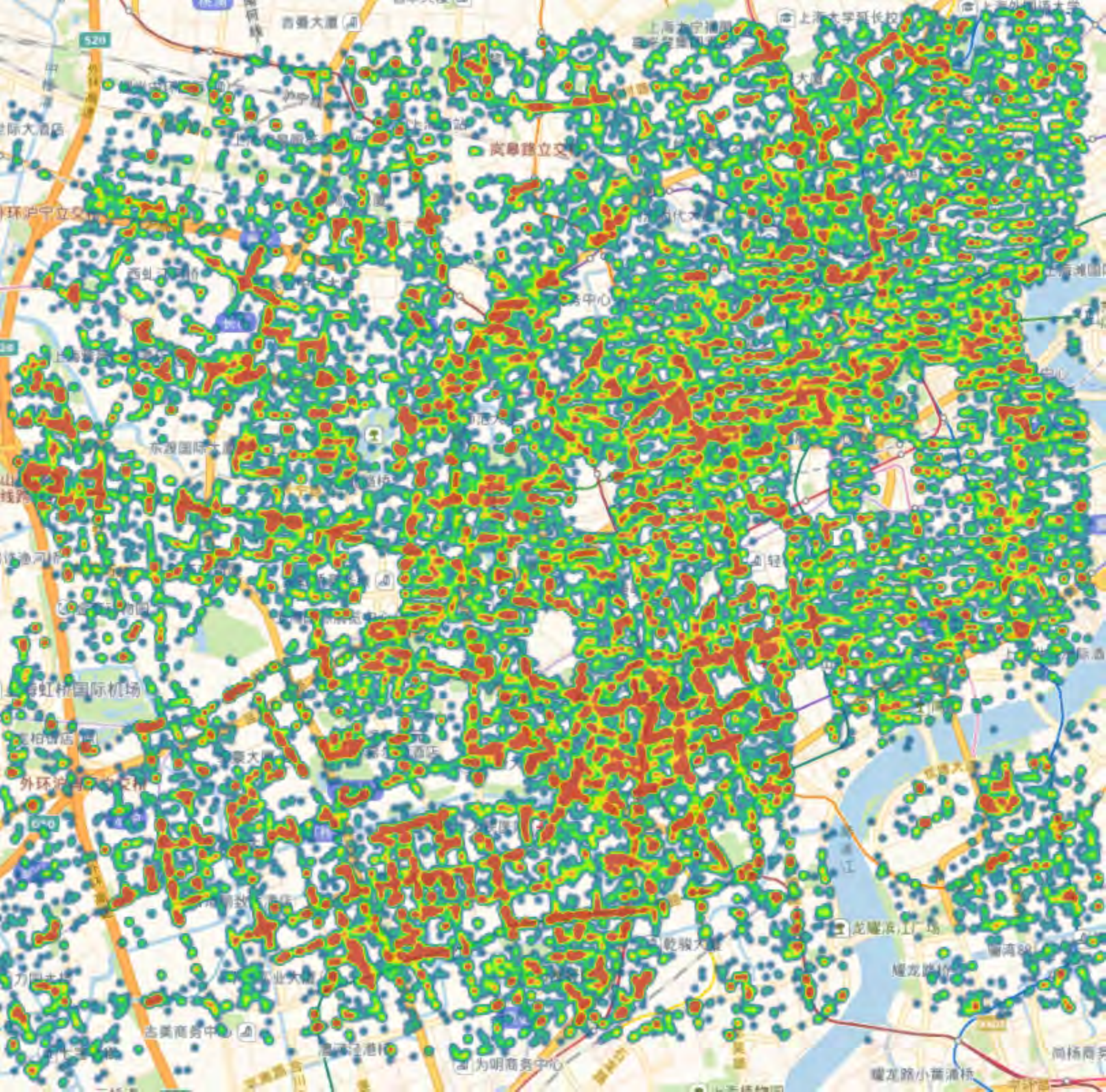}} 
	\subfigure[Pick-up, Weekend]{
		\includegraphics[height=1.3in,width=1.4in, bb=0 0 810 800]{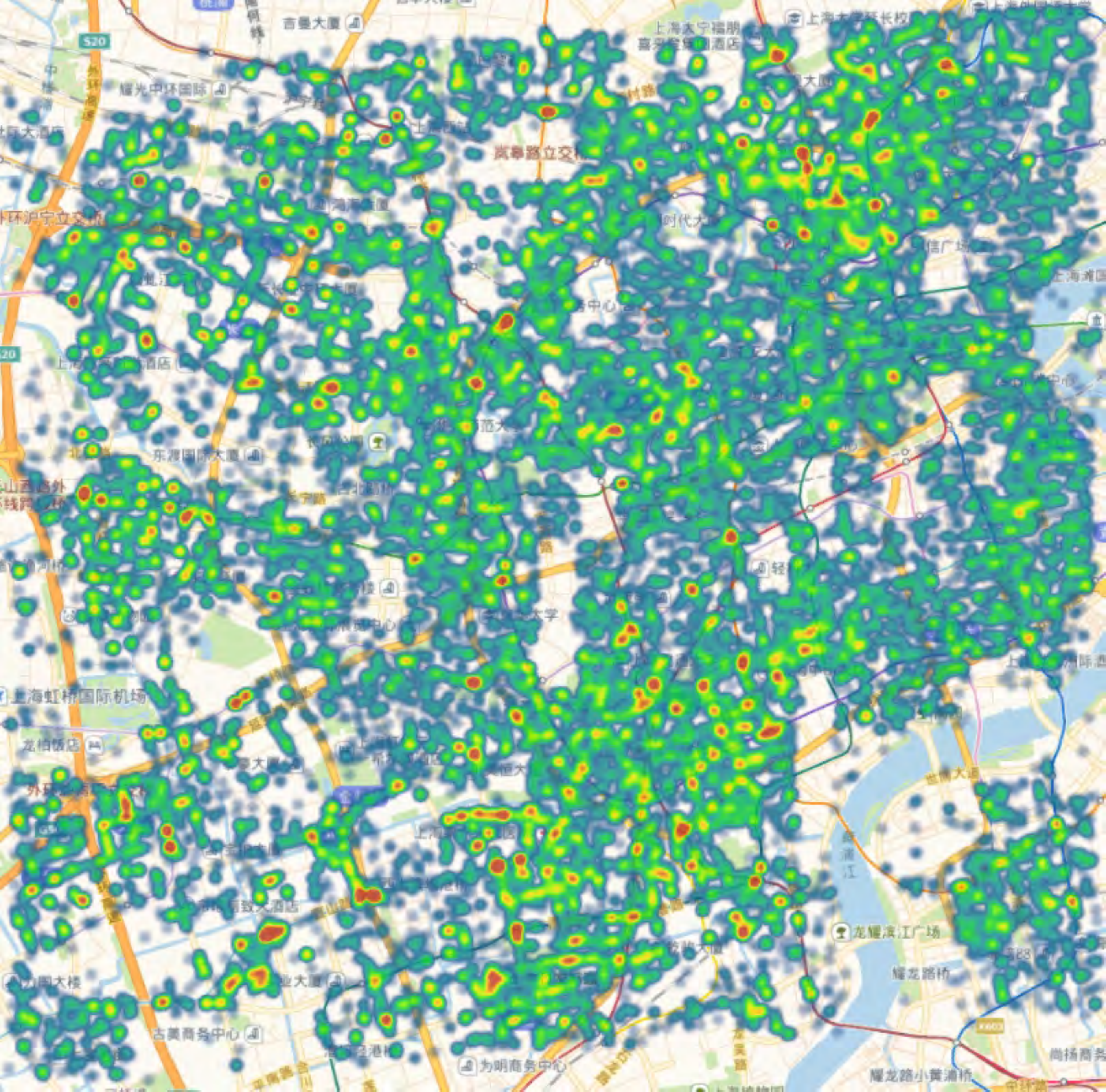}} 
	\subfigure[Drop-off, Weekend]{
		\includegraphics[height=1.3in,width=1.4in, bb=0 0 810 800]{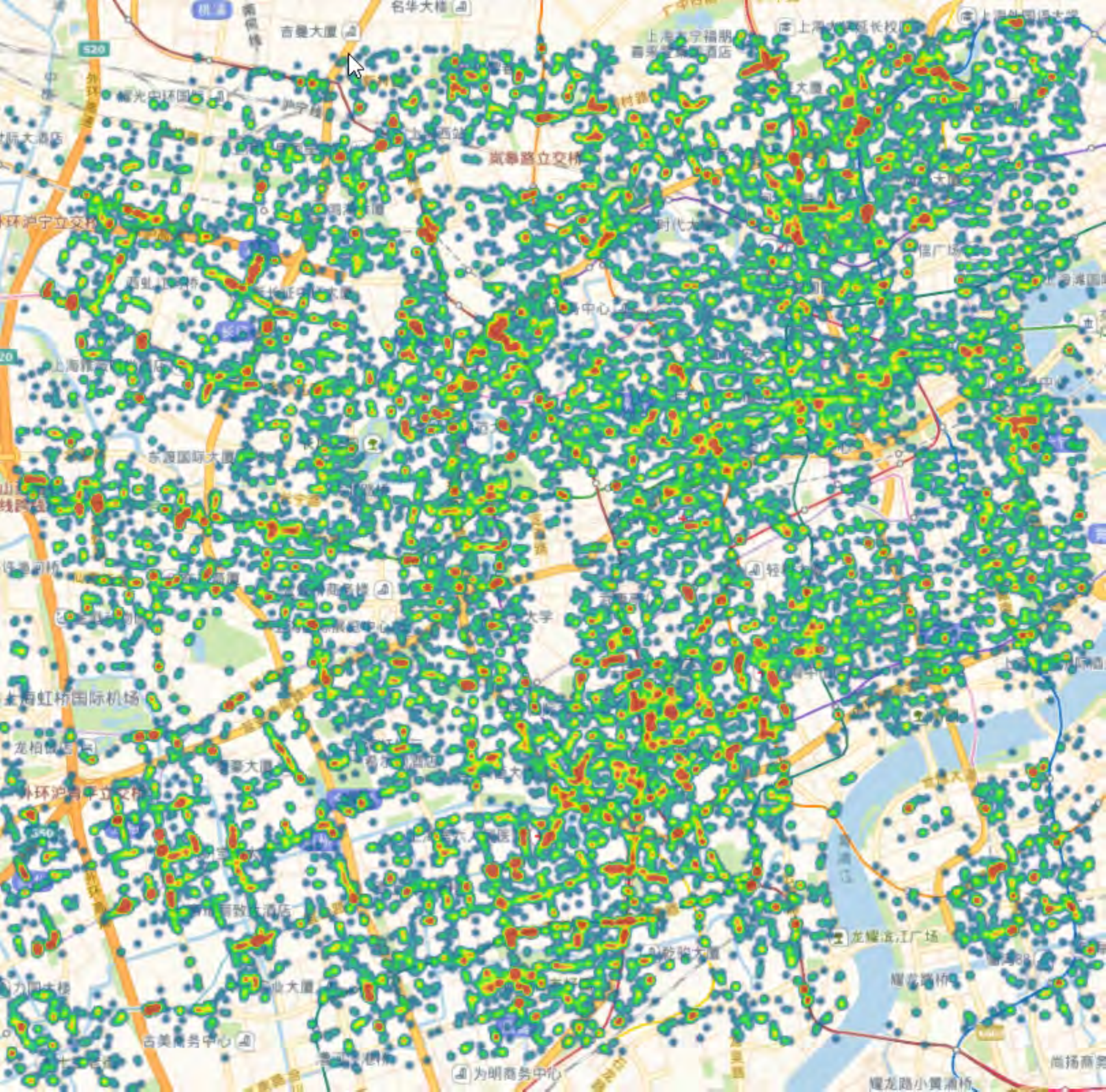}} 
	\vspace{-2mm}
  	\caption{Density distribution.}
	\vspace{-3mm}
    \label{case:Density}
\end{figure*}

\begin{figure*}[t]
	\centering
	\subfigure[Hechuan Road, Working day]{
		\includegraphics[height=1.2in,width=.45\textwidth, bb=0 0 749 329]{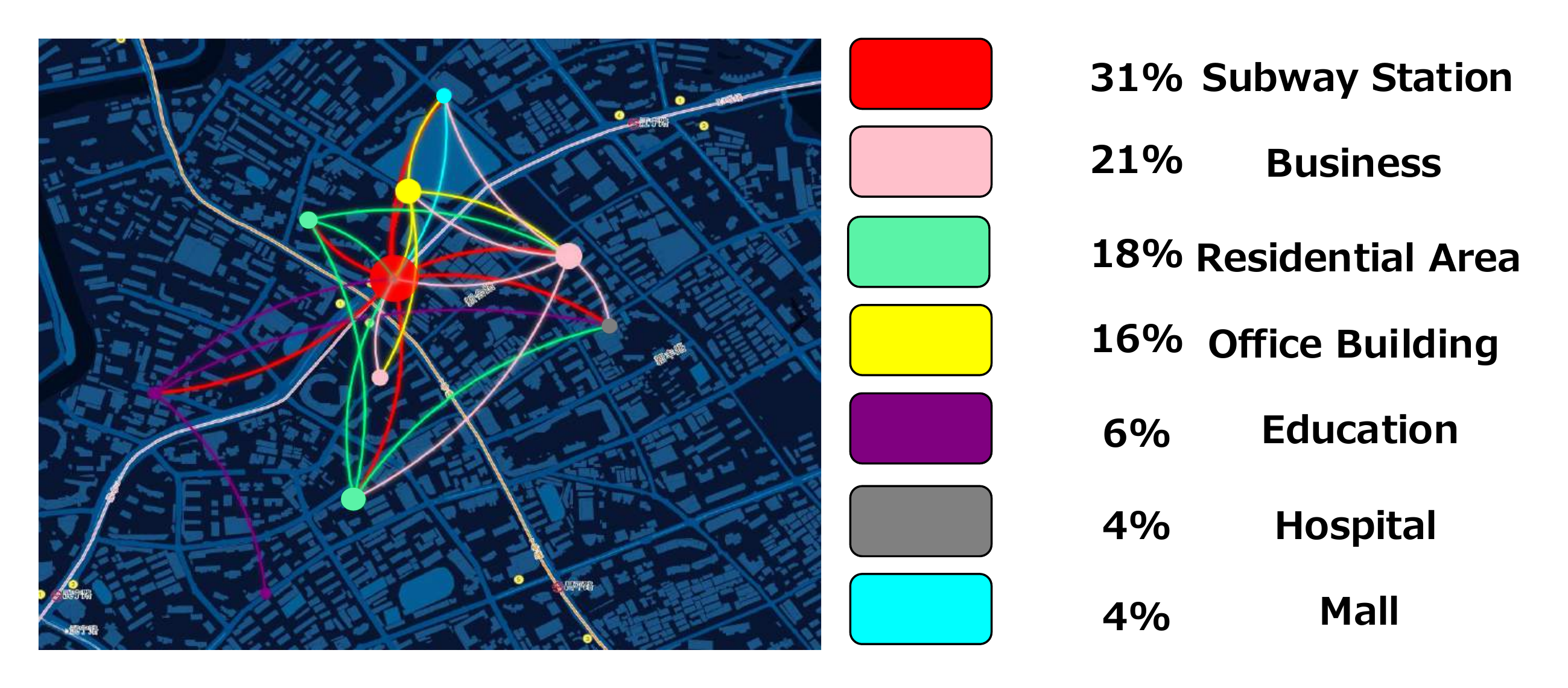}} 
	\subfigure[Changshou Road, Working day]{
		\includegraphics[height=1.2in,width=.45\textwidth, bb=0 0 749 329]{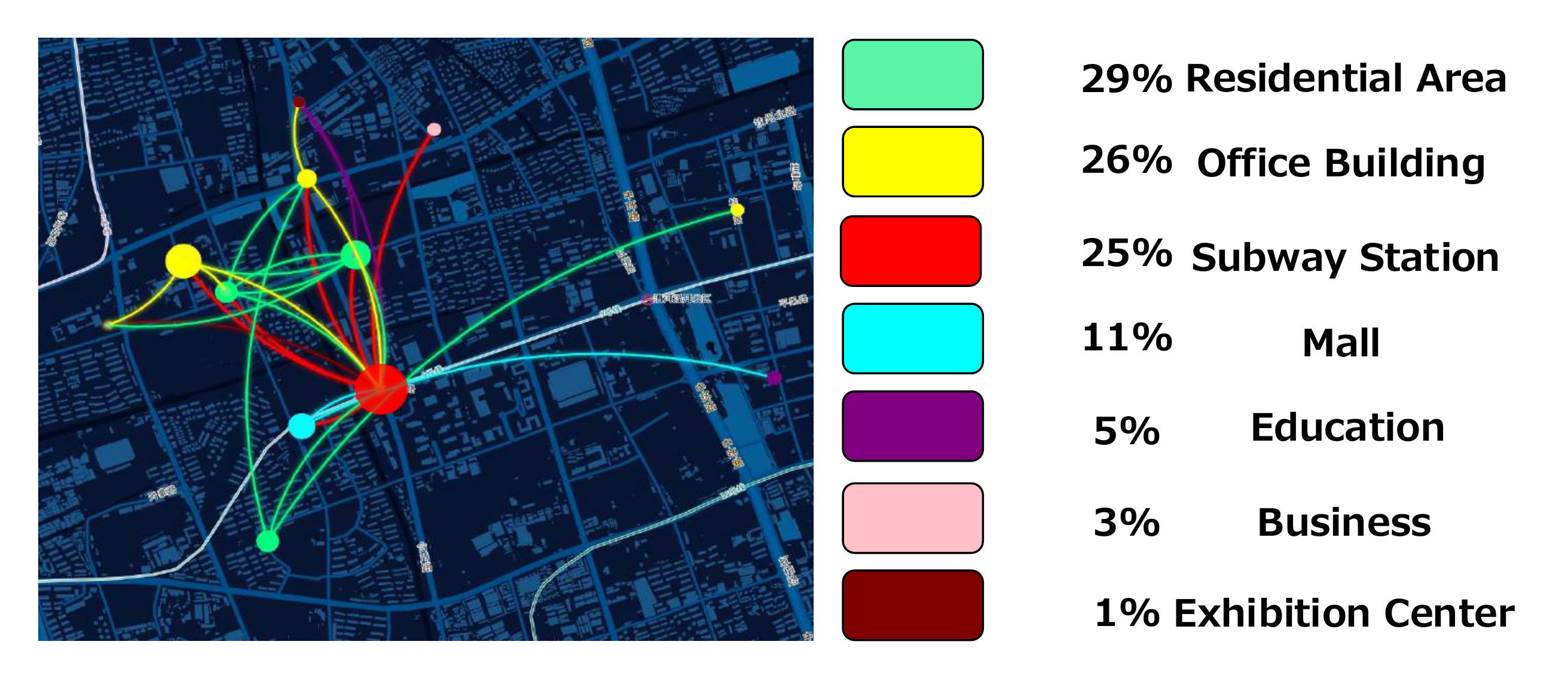}} \\
	\vspace{-2mm}
	\subfigure[Hechuan Road, Weekend]{
		\includegraphics[height=1.2in,width=.45\textwidth, bb=0 0 749 329]{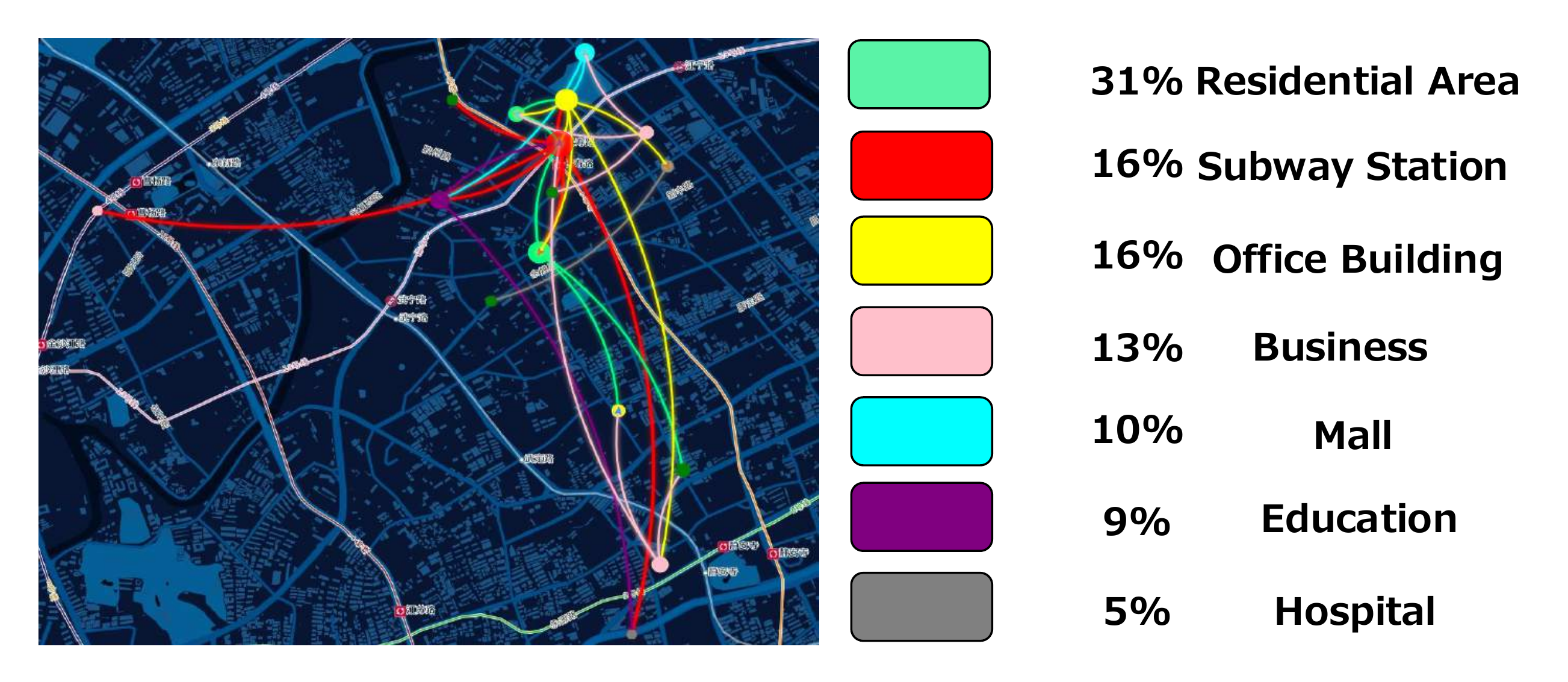}} 
	\subfigure[Changshou Road, Weekend]{
		\includegraphics[height=1.2in,width=.45\textwidth, bb=0 0 749 329]{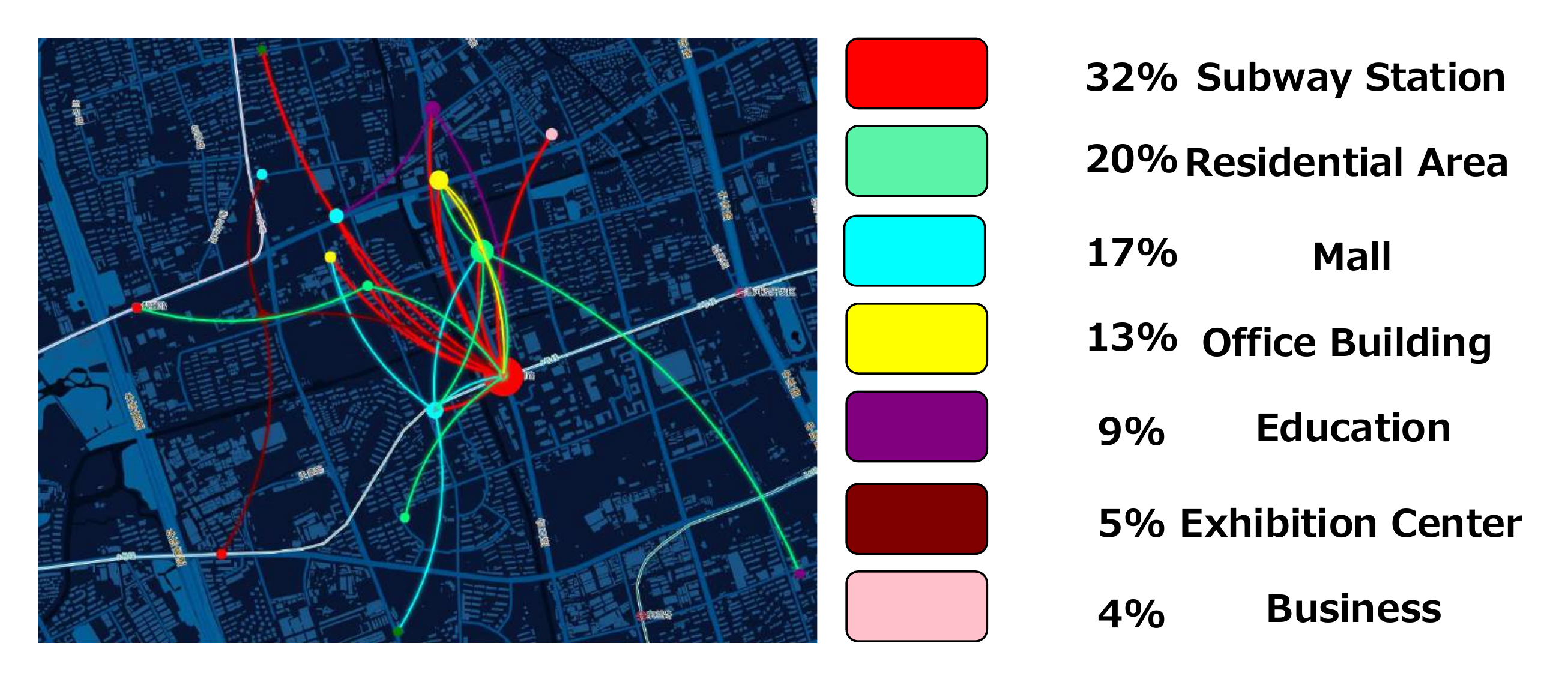}} 
	\vspace{-2mm}
  	\caption{The top flow trajectories and the distribution of hot POIs.}
	\vspace{-4mm}
    \label{case:mobility}
\end{figure*}

Although, considerable research efforts have been paid in urban mobility prediction based on big data, there still lacks research on how to use dockless bike traffic data to probe the user dynamic, while the fixed-dock shared bikes are not flexible enough to analyze the user mobility or dynamic.

In this sub-section, with the help of Point-of-Interest (POI) data, we demonstrate how to use the predicted bike flow to sense the POI distribution and further expose the users trajectory and urban lives. Specifically, the bike data contains the bike ID, travelling time, as well as the location of pick-up and drop-off. Each location can be linked to one or multiple specific POI(s). The results are presented in two categories: (1) The density distribution of shared bikes's pick-up and drop-off in a regular working day and weekend, respectively (\autoref{case:Density} ). (2) The top flow trajectories of shared bikes and the distribution of hot POIs (\autoref{case:mobility}).

\autoref{case:Density} (a) and (b) illustrate the geographical distribution of shared bikes' pick-up and drop-off respectively during a regular working day time (8:00am-12:00am), where the dots colored in red and blue denote high and low bike densities respectively. Then, the \autoref{case:Density} (c) and (d) show the distribution in a weekend day, from which, we can observe that the pick-up of shared bikes distribute evenly in the city, while the distribution of drop-off tends to concentrate in some agglomerated regions. In addition, the total amount of shared bikes' usage in the working day are much higher than those in the weekend, both in the pick-up and drop-off. 

\autoref{case:mobility} shows the distribution of POIs with top 10 travel flows, where each circle presents one POI. The larger the circle, the greater the total flow (including both outflow and inflow). One POI contains several regions segmented following the method proposed in the Section 2. The different color lines represent the top flow trajectories from different categories of POIs. For example, \autoref{case:mobility} (a) illustrates the zone surrounding the Hechuan Road (a famous downtown area) in the working day, which is the same time as the \autoref{case:Density} (a) and (b). From this sub-figure, we can observe that the POI with highest flow is the subway station, which is a transportation transfer and attracts 31$\%$ flows. The second tier of POIs of outflow are business, residential area, office building, hospital, and education (including some schools). 

\autoref{case:mobility} (b) illustrates the data analysis result concurrent with the \autoref{case:mobility} (a). The difference is it in the zone around the Changshou Road with more residents. Thus the POI with highest flow here is the residential area. \autoref{case:mobility} (c) and (d) show the analysis results of above zones at the weekend. Compared with (a) and (b), the circles of POI are smaller, i.e. the total amount of bike flows is less than that in the working day. Moreover, the trajectory distances are much longer, especially in the \autoref{case:mobility} (c), where some trajectories are even twice as long as those in the \autoref{case:mobility} (a). The flow ratios of mall (including restaurant, shopping and cinema ) are much higher.

From this case study, we can find the prediction of shared bike flow has a significant value for sensing and analyzing users’ attentions and activities, as well as the POIs distribution e.g. recreational business districts, transportation hubs, and city landmarks.

\fi
\vspace{-3mm}
\section{Related Work}

Traffic flow prediction is a long-standing problem in urban computing, and time series analysis methods or spatio-temporal correlation have been extensively studied for this task  \cite{zhang2017deep}. In addition, multi-source data such as GPS trajectories, map data and weather conditions have been exploited to produce more robust models \cite{liu2016rebalancing,hulot2018towards}. But these existing studies are mainly for vehicle or crowd flow prediction. Recently, bike sharing systems have attracted increasingly attention due to the wild spread of sharing economy\cite{wang2018bravo}. Many methods, such as regression model \cite{singhvi2015predicting} and Gradient Boosting Tree \cite{li2015traffic}, have been explored for predicting the bike flow. Auxiliary techniques have also been exploited to enhance flow prediction, including multi-source data analysis \cite{liu2018inferring} and clustering \cite{chen2016dynamic,liu2017functional,liu2016rebalancing}. Other tries include utilizing deep learning frameworks to predict flow throughout a spatio-temporal network~\cite{ai2018deep,zhang2019flow,li2018dynamic}. However, these approaches, flaw in the interpretablity extract the hidden traffic patterns. In addition, most of them initially proposed for BSSs with docking stations, which hard to be applied universally in dockless systems.Recently, some works for the traffic prediction have been proposed based on the (Graph) Convolution Network \cite{lin2018predicting,geng2019spatiotemporal,chai2018bike,cui2019traffic,zhang2017deep}, which have excellent performances on flow prediction. However, these (graph) convolution-based models are more appropriate for predicting the overall region-level total flow (only inbound and outbound flow). For the pairwise flow (region to region) prediction, the conv-based methods are too complex in training phase, as they require independent training for each pair of OD (Origin-Destination). There are a few works about OD matrix prediciont \cite{deng2016latent,wang2019origin,gong2019potential}, but the scenarios they considered are not closely relevant to ours.

\vspace{-5mm}
\section{Conclusion}
In this paper, we developed an interpretable bike flow prediction (IBFP) method, which can provide effective bike flow prediction with interpretable flow patterns. Specifically, we first divided the entire city into regions according to the flow density.
Then, we extracted interpretable patterns by subspace clustering with sparse representation.
Next, we modeled the spatio-temporal interactions between regions with a graph regularized sparse representation method, and characterized the  commonalities of periodic data structure by graph smoothing.
After that, we constructed interpretable base matrices from traffic patterns and learned the coefficients.
Finally, experimental results on real-world dockless bike sharing data demonstrated that the proposed IBFP outperformed state-of-the-art methods in terms of prediction accuracy and interpretability. 
In particular, to ensure better generalization, we propose a purely data-driven model which extracts bike flow patterns from historical flow data. However, external features like weather data can be readily incorporated into our model: we can manually form a base matrix with flow data related a specific weather.
\vspace{-3mm}

\bibliographystyle{IEEEtran}
\bibliography{Bibliography}
\vspace{10mm}

\vspace{5mm}
\textcolor{blue}{
\textsf{\textbf{{\Large A}}}\textsf{\textbf{{\large PPENDIX}}}}
\vspace{4mm}

\textbf{A. Joint probability factor model}
\vspace{3mm}

\textbf{(1) Reformulating the optimization problem by a joint probability factor model.}

\begin{figure}[!th]
 	\centering
   	\subfigure[Weekdays flow distribution]{
 		\includegraphics[height=1in,width=0.23\textwidth, bb=0 0 576 432]{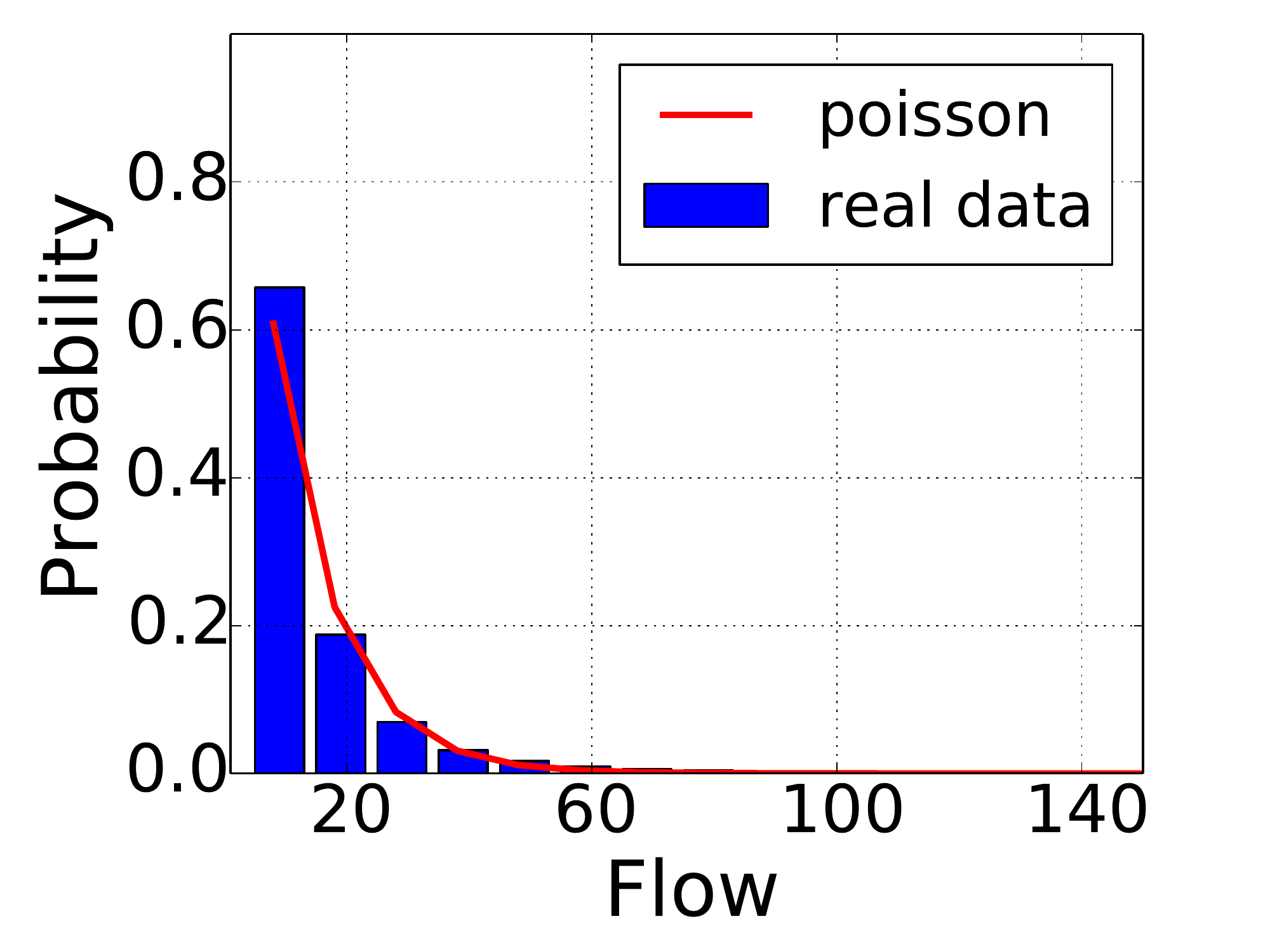}}
        \hspace{-4ex}
 	\subfigure[Holiday flow distribution]{
 		\includegraphics[height=1in,width=0.23\textwidth, bb=0 0 576 432]{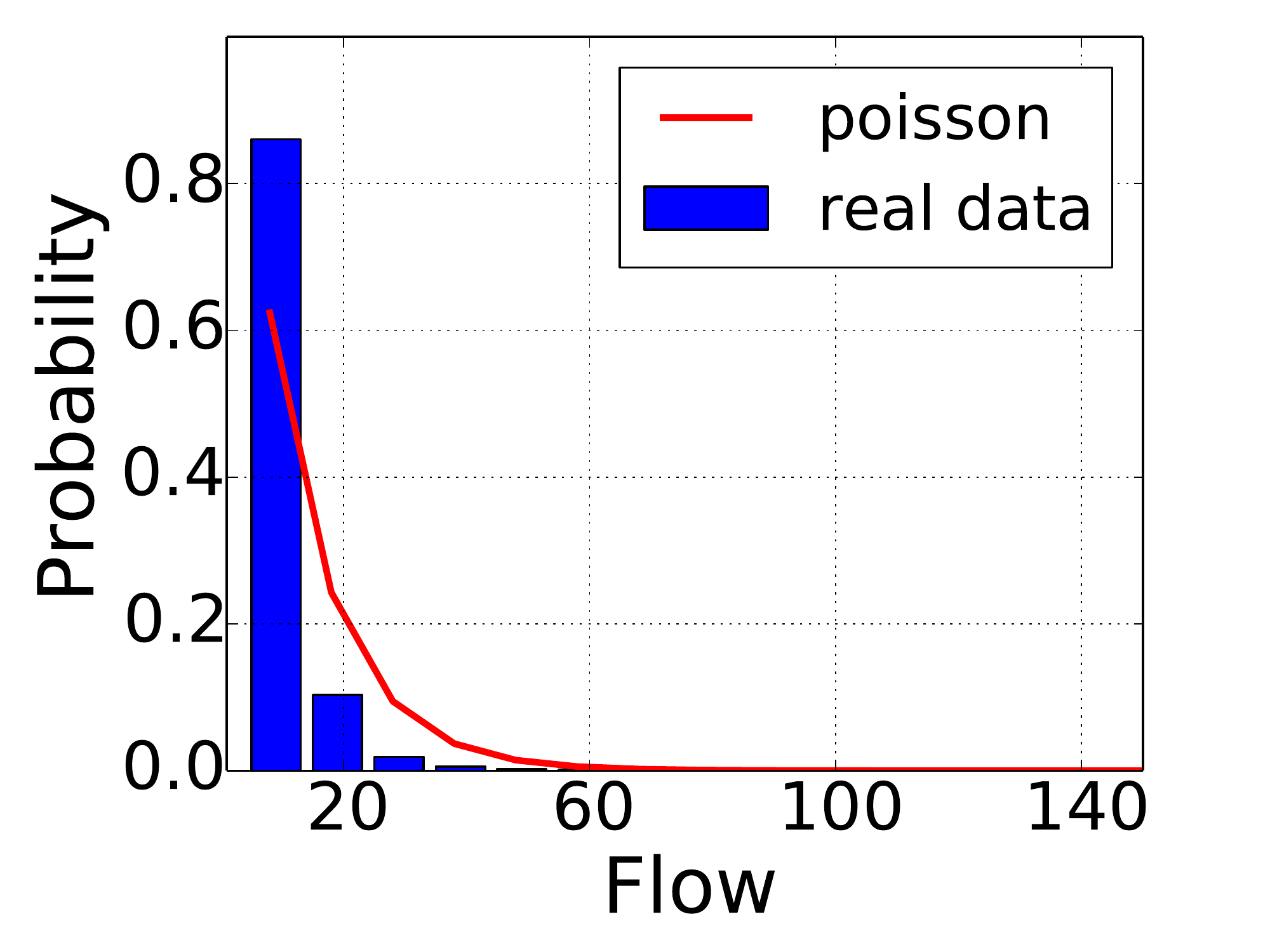}}
 	\vspace{-2mm}
 	\caption{Statistics of flow distribution}
 	\vspace{-3mm}
     \label{fig:statsforprob}
 \end{figure}

The optimization problem \ref{eqn:06} can be shortened as:
\begin{equation}
\label{eqn:07}
\min\mathcal{L}(B,S)+\lambda\cdot\mathcal{K}(S)+\gamma\cdot\Omega(S),
\end{equation}
where $\mathcal{L}(B,S)$ is the loss function $\frac{1}{2}\sum_{i=1}^{N}\Vert F^i-\sum_{c=1}^{C}S_{ic}B^c\Vert_F^2\,$, $\mathcal{K}(S)$ is the lasso term $\sum_{i=1}^{N}\sum_{c=1}^{C}\Vert S_{ic}\Vert_1$, and $\Omega(S)$ represents the graph regularization $tr(S^TLS)$.
We now discuss how to learn the coefficient matrix $S$ of problem in \autoref{eqn:06} with the constructed base matrices $B$. 
Generally, matrix reconstruction methods are mostly based on the assumption that the data is following the Gaussian distribution \cite{huang2018learning}, while this assumption does not hold for the bike flow data (from \autoref{fig:statsforprob}). 
Moreover, incomplete and biased observations \cite{mnih2008probabilistic} which may cause problems for accurately learning the above model. To solve this problem, we first propose a joint probability factor model to approximate the matrix reconstruction optimization base on the distribution of real flow data. Then we learn the coefficients. 

\autoref{fig:statsforprob} shows distributions of the average bike flow count per hour for each pair region in weekdays and weekends, respectively. From them, we find that Poisson distribution can approximate these flow distributions well and also provide non-negative responses. 
Therefore, each flow record $f_{ij}^n$ from the $i$-th to the $j$-th region at the $N$-th time fragment can be approximated by a Poisson distribution with a base element $B_{ij}^*$ :

\begin{equation}
	\label{eqn:10}
	f_{jl}^i\sim Poisson(\langle S_{i*}\,,\,B_{jl}^*\rangle)\,,\quad\forall i,j,l	
\end{equation}
where $\langle S_{i*},B_{jl}^*\rangle=\sum_c S_{ic}B^c$. Basis $B^c$ denotes some attribute interaction pattern of bike flow which also follows the Poisson distribution. Furthermore, we can use the Gamma distribution as a prior since it is the conjugate one of Poisson. $S$ is a coefficient matrix which generally follows the Gaussian distribution \cite{yang2017unified}. Then we have the following generative process:
\begin{equation}
	\begin{split}
		& 1.\quad S_{ic}\sim Gaussian(0\,,\,\sigma)\,,\quad     \forall i,c \\
		& 2.\quad B_{jl}^c\sim Gamma(\eta\,,\,\theta)\,,\quad  \forall c,j,l
	\end{split}
\end{equation}
where $\sigma$ is the parameter of Gaussian distribution, $\eta$ and $\theta$ are the parameters of Gamma distribution. With the above formulation, we have the joint probability density:
\begin{equation}\label{eqn:11}
	\begin{split}
	Pr(F|S,B)&Pr(S)Pr(B)\\
	&=\prod_{i,j,l}{(\frac{(\langle S_{i*},B_{jl}^*\rangle)^{F_{jl}^i}}{\Gamma(F_{jl}^i+1)}\exp(-\langle S_{i*},B_{jl}^*))^{I_{jl}^i}} \\
	& \times \prod_{i,c}\frac{1}{\sqrt{2\pi}\sigma_A}\exp(-\frac{(S_{ic})^2}{2\sigma_A^2}) \\
	& \times \prod_{c,j,l}\frac{\theta^\eta}{\Gamma(\eta)}(B_{jl}^c)^{\eta-1}\exp(-\theta B_{jl}^c),
	\end{split}
\end{equation}
where $\Gamma(\cdot)$ is the gamma function $\Gamma(n+1)=n!$, and $I_{jl}^i=1$ if and only if $F_{jl}^i>0$, otherwise $I_{jl}^i=1$. For much simpler calculation, we re-formulate the loss function  \autoref{eqn:07} in \autoref{eqn:11} to \autoref{eqn:12} by taking the (negative) log-likelihood.
\begin{equation}\label{eqn:12}
	\begin{split}
	\mathcal{L}(B,S) & =-\log Pr(F|S,B)Pr(S)Pr(B)+const \\
	& =-\sum_{i=1}^{N}\sum_{j=1}^{M}\sum_{l=1}^{M}I_{jl}^i(F_{jl}^i\ln\langle S_{i*},B_{jl}^*\rangle\\
	&-\langle S_{i*},B_{jl}^*\rangle)+\frac{1}{2\sigma_S^2}\Vert S\Vert^2 \\
	& -\sum_{c=1}^{C}\sum_{j=1}^{M}\sum_{l=1}^{M}((\eta-1)\log B_{jl}^c-\theta B_{jl}^c).
	\end{split}
\end{equation}

Consequently, The reconstruction error $\mathcal{L}(B,S)$ can be rewritten as $\mathcal{L}(S)$. 

\vspace{4mm}

\textbf{(2) Solving the optimization problem.}
\vspace{3mm}

\rev{Because a $\ell_1$-$norm$ in $lasso$ is non-differentiable at the origin}, the problem of \autoref{eqn:07} can not be solved by the standard Gradient method. Instead, we introduce Proximal Gradient (PG) method to solve this problem ~\cite{chen2012smoothing}.

First, we consider the problem without $\ell_1$-$norm$ and mark $f(S)=\mathcal{L}(S)+\gamma\cdot\Omega(S)$. Suppose $S_i$ is the $i$-th line of the coefficient matrix $S\,$, which is related to the bike flow of the $i$-th time fragment. $(S_{i})_t$ and $(S_i)_{t+1}$ are the values after $t$-th times and $(t+1)$-th times iterations of $S_i$, respectively. Thus, by the second order Taylor expansion of $f(S_i)$ around $(S_i)_t$, we have:
\begin{equation}\label{eqn:13}
	\begin{split}
	f(S_i) & \cong f((S_i)_t)+\langle\nabla f((S_i)_t),S_n-(S_i)_t\rangle\\
	& +\frac{L}{2}\Vert S_i-(S_i)_t\Vert^2 \\
	& =\frac{L}{2}\Vert S_i-((S_i)_t-\frac{1}{L}\nabla f((S_i)_t))\Vert_2^2+const,
	\end{split}
\end{equation}
where $L>0$ is a constant which satisfies L-Lipschitz constraint, $const$ is a constant independent of $S_i\,$. As a result, the problem of learning the $S_i$ to minimize $f(S_i)$ can be shortened as:
\begin{equation}\label{eqn:14}
	\arg\min\limits_{S_{ic}}\frac{L}{2}\Vert S_i-((S_i)_t-\frac{1}{L}\nabla f((S_i)_t))\Vert_2^2,
\end{equation}
Thus, combing \autoref{eqn:07}, we can obtain $(S_i)_{t+1}$ as follows:
\begin{equation}\label{eqn:15}
	\begin{split}
	(S_i)_{t+1} & =\arg\min\limits_{S_{i_0}}+\lambda\Vert S_i\Vert_1 \\
	& =\arg\min\limits_{S_i}\frac{L}{2}\Vert S_i-((S_i)_t\\
	& -\frac{1}{L}\nabla f((S_i)_t))\Vert_2^2+\lambda\Vert S_i\Vert_1. 
	\end{split}
\end{equation}
For each element, we can obtain:	
 \begin{equation}\label{eqn:16}
 \begin{split}
 	(S_{ic})_{t+1}&=\arg\min\limits_{S_{ic}}(\frac{L}{2}(S_{ic}((S_{ic})_t\\
 	&-\frac{1}{L}\nabla f((S_{ic})_t)))^2+\lambda|S_{ic}| 
 \end{split}
 \end{equation}
Now, the solution process of complex optimal problem in \autoref{eqn:07} without Lasso regularization can be replaced by those of the problem of  \autoref{eqn:15}. Combine \autoref{eqn:15} with the probabilistic approximation in \autoref{eqn:12}, the element-wise gradients are given as below:
\begin{equation}\label{eqn:17}
	\begin{split}
	&\frac{\partial(\mathcal{L}(S)+\gamma\cdot\Omega(S))}{\partial S_{ic}}\\
	& =\frac{\partial(f(S_i))}{\partial S_{ic}} \\
	& =-\sum_{j,l=1}^{M}I_{jl}^n\cdot(\frac{F_{jl}^i\cdot B_{jl}^c}{\langle A_{c,*},B_{jl}^c\rangle}-B_{jl}^c) 
	+\frac{1}{\sigma_S^2}S_{ic} \\
	& +\sum_{i=1}^{N}\sum_{j=1}^{M}\sum_{l=1}^{M}I_{jl}^i\cdot(F_{jl}^i-\sum_{c=1}^{C}A_{jl}\cdot B_{jl}^c)\cdot(-B_{jl}^c) \\
	& +\gamma\frac{\partial tr(S^TLS)}{\partial S_{ic}}.
	\end{split}
\end{equation}
Let $z=(S_i)_t-\frac{1}{L}\nabla f((S_i)_t)$, from the solution of proximal gradient method, we can obtain:
\begin{equation}
	(S_{ic})_{t+1}=\begin{cases}
	z^c-\frac{\lambda}{L},& \frac{\lambda}{L} <z^c \\
	0,                 & |z^c|\leq\frac{\lambda}{L} \\
	z^c+\frac{\lambda}{L},& z^c<-\frac{\lambda}{L}.
	\end{cases}
\end{equation}

\end{document}